\documentclass[lettersize,journal]{IEEEtran}
\usepackage{amsmath,amsfonts}
\usepackage{algorithmic}
\usepackage{algorithm}
\usepackage{array}
\usepackage[font=small]{subfig}
\usepackage{textcomp}
\usepackage{stfloats}
\usepackage{xurl}
\usepackage{verbatim}
\usepackage{graphicx}
\usepackage{cite}
\def\BibTeX{{\rm B\kern-.05em{\sc i\kern-.025em b}\kern-.08em
    T\kern-.1667em\lower.7ex\hbox{E}\kern-.125emX}}
\usepackage{balance}
\usepackage{xcolor}         % colors
\usepackage{acronym}
\usepackage{psfrag}
\usepackage{adjustbox}
\usepackage{booktabs} 
\usepackage{amssymb}
\usepackage{mathtools}
\usepackage{amsthm}
\usepackage{hyperref}
\usepackage{breakurl}
\usepackage{multirow}
 \usepackage{zref-base}
\usepackage{atveryend}
  \makeatletter
\AfterLastShipout{%
  \if@filesw   
    \begingroup
      \zref@setcurrent{default}{\the\value{equation}}%
      \zref@wrapper@immediate{%
        \zref@labelbyprops{eq-last}{default}%
      }%
            \zref@setcurrent{default}{\the\value{figure}}%
      \zref@wrapper@immediate{%
        \zref@labelbyprops{fig-last}{default}%
      }%
    \endgroup
  \fi
}
\makeatother

 \usepackage{xr}
\usepackage{zref-xr}
\theoremstyle{definition}
\newtheorem{theorem}{Theorem}

\newtheorem{lemma}[theorem]{Lemma}

\theoremstyle{definition}

\theoremstyle{remark}

\newcommand\V[1]  { \mathbf{#1} }
\newcommand\B[1]  { \boldsymbol{#1} }

\newcommand\set[1] {\mathcal{#1}}
\newcommand\up[1] {\mathrm{#1}}

\definecolor{green}{RGB}{23, 185, 100}
\definecolor{myblue}{RGB}{112, 117, 179}
\definecolor{myorange}{RGB}{217, 95, 2}
\makeatletter
\newcommand{\acposs}[1]{%
 \expandafter\ifx\csname AC@\AC@prefix#1\endcsname\AC@used
   \acs{#1}'s%
 \else
   \aclu{#1}'s (\acs{#1})%
 \fi
}
\makeatother

\acrodef{awa}[AWA]{`animals with attributes'}
\acrodef{LF}[LF]{labeling function}
\acrodef{MRC}[MRC]{minimax risk classifier}
\acrodef{WMRC}[MMP]{minimax prediction}
\acrodef{PWS}[PWS]{programmatic weak supervision}
\acrodef{MV}[MV]{majority vote}
\acrodef{BF}[BF]{Balsubramani-Freund}
\acrodef{DS}[DS]{Dawid-Skene}
\acrodef{EBCC}[EBCC]{enhanced Bayesian classifier combination}
\acrodef{AMCL}[AMCL]{adversarial multi class learning}

\newcommand{\paperTitleMarkboth}{Reliable Programmatic Weak Supervision with Confidence Intervals for Label Probabilities}

\begin{document}

\title{Reliable Programmatic Weak Supervision with\\Confidence Intervals for Label Probabilities}

\author{
	\vspace{0.2cm}
	\IEEEauthorblockN{Ver\'{o}nica \'{A}lvarez, %~\IEEEmembership{Member,~IEEE,} 
		Santiago~Mazuelas,~\IEEEmembership{Senior~Member,~IEEE,}\\
  Steven~An,\IEEEmembership{}
  and Sanjoy~Dasgupta~\IEEEmembership{}} 	
        % <-this % stops a space
        \thanks{Manuscript received September 30, 2024; accepted August 4, 2025.}%
\thanks{Funding in direct support of this work has been provided by projects PID2022-137063NBI00 and CEX2021-001142-S funded by MCIN/AEI/10.13039/501100011033 and the European Union “NextGenerationEU”/PRTR, program BERC-2022-2025 funded by the Basque Government, and grant IIS-2211386 from the national science foundation (NSF). Verónica Álvarez holds a postdoctoral grant from the Basque Government.}% <-this % stops a space
\thanks{V.~\'{A}lvarez is with the Massachusetts Institute of Technology (MIT), Cambridge 02139, USA and BCAM-Basque Center for Applied Mathematics, Bilbao 48009, Spain (e-mail: vealvar@mit.edu).}% <-this % stops a space
\thanks{S.~Mazuelas is with the BCAM-Basque Center for Applied Mathematics, and IKERBASQUE-Basque Foundation for Science, Bilbao 48009, Spain (e-mail: smazuelas@bcamath.org).}%
\thanks{S.~An is with the University of California, San Diego, San Diego (CA) 92093, United States (e-mail: sla001@ucsd.edu).}%
\thanks{S.~Dasgupta is with the University of California, San Diego, San Diego (CA) 92093, United States (e-mail: sadasgupta@ucsd.edu).}
}

% \markboth{IEEE Transactions on Pattern Analysis and Machine Intelligence}%{Shell \MakeLowercase{\textit{et al.}}: A Sample Article Using IEEEtran.cls for IEEE Journals}

\maketitle

\markboth{IEEE Transactions on Pattern Analysis and Machine Intelligence}{\'{A}lvarez, Mazuelas, An, and Dasgupta: \paperTitleMarkboth}


\begin{abstract}
The accurate labeling of datasets is often both costly and time-consuming. Given an unlabeled dataset, programmatic weak supervision obtains probabilistic predictions for the \mbox{labels} by leveraging multiple weak \mbox{labeling functions (LFs)} that provide rough guesses for labels. Weak LFs commonly provide guesses with assorted types and unknown interdependences that can result in unreliable predictions. Furthermore, existing techniques for programmatic weak supervision cannot provide assessments for the reliability of the probabilistic predictions for labels. This paper presents a methodology for programmatic weak supervision that can provide confidence intervals for label probabilities and obtain more reliable predictions. In particular, the methods proposed use uncertainty sets of distributions that encapsulate the information provided by LFs with unrestricted behavior and typology. Experiments on multiple benchmark datasets show the improvement of the presented methods over the state-of-the-art and the practicality of the confidence intervals presented.
\end{abstract}


\section{Introduction}

\IEEEPARstart{F}{or} many machine learning applications, the accurate labeling of datasets is both costly and time-consuming~\mbox{\cite{li2022selective, fatras2021wasserstein, zhou2023asymmetric, li2024transferring}}. Given an unlabeled dataset, methods for programmatic weak supervision aim to leverage multiple weak \acp{LF} to provide accurate labels \cite{wulearning, ratner2016data}. Since common \acp{LF} only provide rough guesses for labels, programmatic weak supervision methods use the outputs of multiple \acp{LF} to obtain probabilistic predictions for the label of each instance \cite{Ratner2017, shin2021universalizing, Fu2020, Mazzetto2021, vishwakarma2022lifting, kuang2022firebolt, tonolini2023robust}. These predictions can then be used to create a fully supervised dataset composed by the instances corresponding to high-confidence predictions, e.g., a label with a large enough predicted probability is regarded as the actual label. Hence, in cases where the probabilistic predictions are reliable, programmatic weak supervision methods can provide high-quality training data and reduce the need for meticulous and expensive labeling of datasets \cite{Ratner2017, zhang2023leveraging, wulearning}.

Most of existing methods for programmatic weak supervision obtain probabilistic predictions by using specific statistical models for the relationship between \acp{LF}' guesses and actual labels. In particular, some works assume that the multiple guesses are independent given the actual label \cite{dawid1979maximum}, while other methods model additional dependences through more general graphical models~\mbox{\cite{Fu2020, li2019exploiting, varma2019learning, bach2017learning}}. The existing methods that do not rely on such dependence modeling utilize a minimax formulation based on estimates for the error of \acp{LF} that provide label guesses~\mbox{\cite{Mazzetto2021, arachie2021general, balsubramani2016optimal}}.

\begin{figure*}
         \centering
         \psfrag{f2}[][][1]{\textbf{Unknown labels}}
         \psfrag{f1}[][][1]{\textbf{Unlabeled instances}}
         \psfrag{f3}[][][1]{\textbf{Weak \acp{LF}}}
         \psfrag{f5}[][][1]{\textbf{Guesses:}}
         \psfrag{f6}[][][1]{ }
         \psfrag{f4}[][][1]{\textbf{Proposed methodology}}
         \psfrag{R}[][][1]{Review:}
         \psfrag{R1}[][][1]{\textcolor{myblue}{Positive (POS)}}
         \psfrag{R2}[][][1]{\textcolor{myorange}{Negative (NEG)}}
         \psfrag{LFs}[][][1]{LFs: }
         \psfrag{ea}[][][1]{$L^3$: bad, $L^4$: hate}
         \psfrag{s}[][][1]{$\Rightarrow$}
        \psfrag{eo}[][][1]{\textcolor{myorange}{\text{NEG}}}
        \psfrag{sa}[][][1]{$L^1$: like, $L^2$: love}
         \psfrag{e}[][][1]{$\Rightarrow$}
         \psfrag{so}[][][1]{\textcolor{myblue}{\text{POS}}}
         \psfrag{da}[][][1]{$L^5$: but}
         \psfrag{d}[][][1]{$\Rightarrow$}
         \psfrag{do}[][][1]{ }
         \psfrag{p}[][][1]{Probabilistic predictions}
         \psfrag{o}[][][1]{and}
         \psfrag{r}[][][1]{prediction}
         \psfrag{j}[c][][1]{predictions and}
        \psfrag{a}[][][1]{confidence intervals}
         \psfrag{c}[][][1]{distributions}
         \psfrag{i}[][][1]{interval}
         \psfrag{w}[][][1]{}
         \psfrag{r1}[][][1]{``\,The movie is not that \textcolor{myorange}{bad},}
          \psfrag{r2}[][][1]{\textcolor{myorange}{but} I \textcolor{myorange}{hate} when Van Damme}
          \psfrag{r3}[][][1]{has \textcolor{myblue}{love} in his movies.\,''}
          \psfrag{r4}[][][1]{``\,If you  \textcolor{myblue}{like} original gut}
          \psfrag{r5}[][][1]{wrenching laughter you}
          \psfrag{r6}[][][1]{will \textcolor{myblue}{like} this movie.\,''}
         \psfrag{1}[][][1]{1}
         \psfrag{2}[][][1]{2}
         \psfrag{LF1}[][][1]{0.6, \textcolor{myblue}{\text{POS}}, \textcolor{myorange}{\text{NEG}}, \textcolor{myorange}{\text{NEG}}, \textcolor{myblue}{\text{POS}}, ?}
         \psfrag{L2}[][][1]{\textcolor{myblue}{\text{POS}}}
         \psfrag{L5}[][][1]{\textcolor{myorange}{\text{NEG}}}
         \psfrag{L4}[][][1]{\textcolor{myorange}{\text{NEG}}}
         \psfrag{L3}[][][1]{\textcolor{myblue}{\text{POS}}}
         \psfrag{L6}[][][1]{?}
         \psfrag{L7}[][][1]{?}
         \psfrag{L8}[][][1]{\textcolor{myblue}{\text{POS}}}
         \psfrag{L9}[][][1]{?}
         \psfrag{Lc}[][][1]{$L^1$}
         \psfrag{Ld}[][][1]{$L^2$}
         \psfrag{Le}[][][1]{$L^3$}
         \psfrag{Lf}[][][1]{$L^4$}
         \psfrag{Lu}[][][1]{$L^5$}
         \psfrag{L0}[][][1]{?}
         \psfrag{La}[][][1]{?}
         \psfrag{Lb}[][][1]{\textcolor{myblue}{\text{POS}}}
         \psfrag{4}[][][1]{0.8}
         \psfrag{0}[][][1]{0.55}
         \psfrag{3}[][][1]{0.5}
         \psfrag{7}[r][r][0.9]{1}
         \psfrag{5}[r][r][0.9]{0.9}
         \psfrag{6}[r][r][0.9]{0.8}
         \psfrag{8}[][b][1]{1}
         \psfrag{9}[][b][1]{2}
         \psfrag{p1}[r][][1]{}
          \psfrag{p2}[r][][1]{}
          % \psfrag{a}[][][1]{Proposed}
          \psfrag{b}[][][1]{method} 
          \psfrag{f5}[][][1]{$\set{U}$} 
          \psfrag{u}[c][][1]{set of} 
          \psfrag{q}[][][1]{Uncertainty} \includegraphics[width=0.95\textwidth]{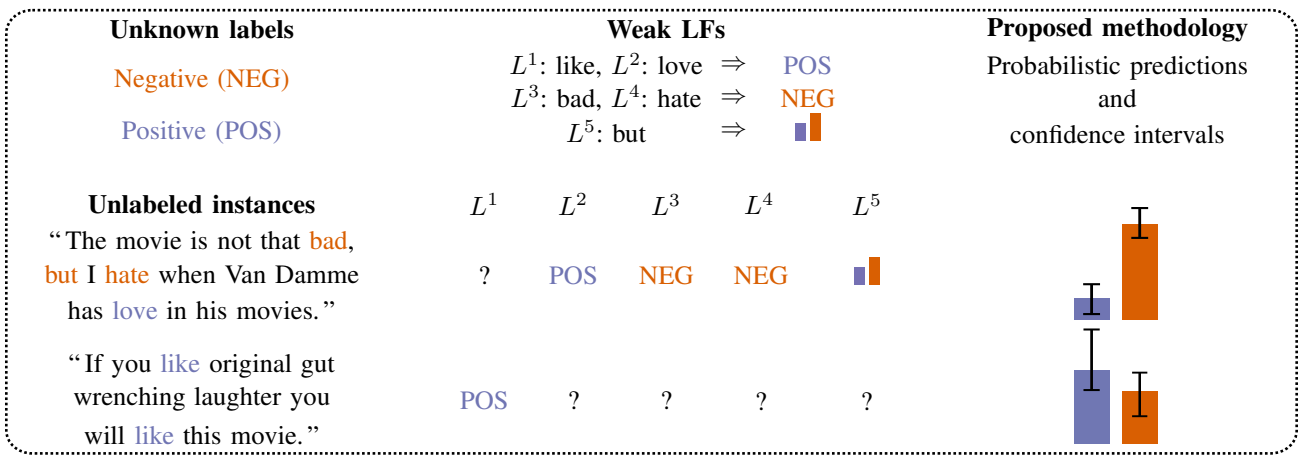}
          % \vspace{-0.25cm}
         \caption{
Example of programmatic weak supervision where the unlabeled dataset is composed by reviews (left column) and the LFs are simple heuristics based on the presence of specific words (middle top). In addition, the LFs provide rough guesses (middle bottom) that have assorted types and unknown interdependences. % The proposed methodology can leverage guesses from general weak LFs and provide reliable probabilistic predictions together with confidence intervals for label probabilities. 
The proposed methodology determines an uncertainty set of distributions that encapsulate the information provided by general LFs. This set contains probability distributions consistent with the LFs' observed behavior such as abstention rate and bounds for the LFs' error.  Then, the proposed methodology uses the uncertainty set to provide reliable probabilistic predictions and confidence intervals.
 }
         \label{fig:intro}
  \end{figure*}

Weak \acp{LF} are inexpensive and often highly unreliable rules
given by heuristics, rules-of-thumb \cite{shin2015incremental}, external knowledge bases \cite{mintz2009distant}, and simple pre-trained models \cite{das2020goggles}. \acp{LF} used in practice commonly provide guesses with assorted types and have largely unknown interdependences \cite{bach2017learning, Zhang2022a, ratner2016data}. Common \acp{LF} not only provide label guesses but can also abstain in certain instances or provide guesses for label probabilities (see Figure~\ref{fig:intro}). In addition, the \acp{LF} often provide guesses that are conflicting and noisy or have an unknown and complex dependence (e.g., \acp{LF} trained with overlapping datasets). 

Existing methods for programmatic weak supervision can provide unreliable probabilistic predictions when the \acp{LF} provide guesses that do not comply with specific statistical behaviors or typologies. These unreliable probabilities can significantly reduce the benefits of programmatic weak supervision. For instance, if the labels with predicted probabilities larger than $0.9$ are regarded as correct but only $60\%$ of them are accurate, classification rules subsequently learnt using such labels would be highly inaccurate. Furthermore, existing techniques for programmatic weak supervision cannot assess the reliability of the probabilistic predictions for labels.  Such assessments can significantly improve programmatic weak supervision systems providing additional levels of information that result in more dependable labeling processes.  For instance, a confidence interval of $[0.5,0.7]$ for the probabilities corresponding to predictions larger than $0.9$ would indicate that the predicted probabilities are significantly unreliable in that range, and that the threshold of $0.9$ is inadequate to regard labels as correct. % Unfortunately, conventional statistical tools can only obtain confidence intervals for label probabilities in rather limited situations, such as those when \acp{LF}' guesses comply with a known and simple statistical model \cite{li2014error}. 
The assessment of probabilistic predictions by means of confidence intervals has been mainly studied for the problem of binary classification~\cite{barber2020distribution, gupta2020distribution}. Specifically, the methods in~\cite{barber2020distribution, gupta2020distribution} do not rely on distributional assumption and provide confidence intervals for label probabilities using a so-called ``calibration set'' formed by labeled samples. However, these methods cannot be used in programmatic weak supervision since they require a sizable number of labeled samples. Existing methods for programmatic weak supervision can only obtain confidence intervals for label probabilities in rather limited situations, such as those when \acp{LF}' guesses comply with a known and simple statistical model~\cite{li2014error}.

This paper presents a methodology for programmatic weak supervision that can use general \acp{LF} to provide confidence intervals for label probabilities together with more reliable probabilistic predictions (see Fig.~\ref{fig:intro}). The proposed methodology determines an uncertainty set of distributions that encapsulates the information provided by general LFs. Given the uncertainty set, the proposed methodology obtains probabilistic predictions by minimizing the worst-case expected log-loss and provides confidence intervals given by the worst-case probabilities over groups of instances. The use of uncertainty sets allows the proposed methods to assess the reliability of probabilistic predictions by considering all the distributions that are consistent with the LFs' observed behavior.  
Specifically, the main contributions of the paper are as follows.

\begin{itemize}
\setlength{\itemsep}{0.5em}
\item We develop programmatic weak supervision methods that use an uncertainty set of distributions to encapsulate the information provided by \acp{LF} with unrestricted behavior and typology.
\item We present methods that provide confidence intervals for the label probabilities corresponding to any group of instances. In addition, the theoretical results presented characterize the validity and optimality of such confidence intervals. 
\item We develop techniques that use general \acp{LF} to obtain probabilistic predictions that minimize worst-case expected losses. In addition, the theoretical results characterize the reliability of such predictions, which are always included in the confidence intervals.
\item We experimentally show the practicality of the confidence intervals presented and the improvement provided by the methods proposed in comparison with existing approaches.
\end{itemize}

The rest of the paper is organized as follows. Section~\ref{sec:WMRC} describes the problem formulation and presents the learning methodology based on uncertainty sets. We propose confidence intervals for label probabilities in Section~\ref{sec:confidence_predictions} and minimax probabilistic predictions in Section~\ref{sec:prob_pred}. Section~\ref{sec:uncertainty_sets} describes feature mappings and expecations estimates. Section~\ref{sec:experimental_results} assesses the proposed methods using benchmark datasets. 

\emph{Notation:} Calligraphic letters represent sets; bold lowercase letters represent vectors; \mbox{$\|\cdot\|_1$ and $\|\cdot\|_\infty$} denote the 1-norm 
and the infinity norm of its argument, respectively; $[\,\cdot\,]^{\top}$ denotes the transpose of its argument; $\preceq$ and $\succeq$ denote vector inequalities; $\mathbb{E}_\up{p}\{\cdot\}$ denotes expectation w.r.t. probability distribution
$\up{p}$; and $\mathbb{I}$ denotes the indicator function. For the readers’
convenience, we provide in Table~\ref{notions} a list with the main notions used in the paper and their
corresponding notations.

\begin{table*}
\centering
\small
\caption{Main notations used in the paper}
\label{notions}
\adjustbox{max width=1.5\textwidth}{%
\begin{tabular}{ll}
\toprule
Notation &Meaning\\
\midrule
\midrule
$y_i \, \in \, \set{Y} = \{1, 2, \ldots, T\}$& label for instance $i \,  \in \, \{1, 2, \ldots, n\}$ \\
$L_i^j$& guess of the $j$-th LF for instance $i$\\
\mbox{$\Lambda_i=[L_i^{1}, L_i^{2}, \ldots, L_i^{m}]$}& array given by the guesses of $m$ LFs for instance $i$\\
$\up{h}_i = [\up{h}_i(1), \up{h}_i(2), \ldots, \up{h}_i(T)]^\top$&probabilistic prediction for instance $i$\\
$\ell(\up{h}_i,y_i)$& loss of probabilistic prediction $\up{h}_i$ with respect to label $y_i$\\
$\set{I}\subseteq\{1, 2, \ldots, n\}$& group of instances\\
$\Delta(\set{Y})$& set of probability distributions over labels $\set{Y}$\\
$\up{p}_\set{I}^*(y)$& actual probability of label $y$ for instances in $\set{I}$\\
$\up{h}_\set{I}(y)$& predicted probability of label $y$ for instances in $\set{I}$\\
$\big[\,\underline{\up{p}}_\set{I}(y)\,,\, \overline{\up{p}}_\set{I}(y)\,\big]$& confidence interval for probability of label $y$ for instances in $\set{I}$\\
$\big[\,\underline{\up{p}}_\set{I}^*(y)\,,\, \overline{\up{p}}_\set{I}^*(y)\,\big]$& optimal confidence interval for probability of label $y$ for instances in $\set{I}$\\
$\Phi(\cdot, \cdot) \, \in \, \mathbb{R}^d$&feature mapping with dimension $d$\\
$\set{U}$&uncertainty set of distributions given by expectations' constraints as in~(1)\\
$\widehat{\B{\tau}} \in \mathbb{R}^d$& estimate of the feature mapping's expectation $\B{\tau}=\frac{1}{n}\sum_{i=1}^n\Phi(\Lambda_i,y_i)$\\
$\B{\lambda} \succeq \V{0}$& assessment of the error in the expectation estimate\\
$R(\set{U})$& minimax risk\\
$\B{\mu}$& vector parameter that determines probabilistic predictions\\
\bottomrule
\end{tabular}}
\end{table*}

\section{Methodology for programmatic weak supervision based on uncertainty sets} \label{sec:WMRC}

This section describes the problem formulation and the proposed uncertainty sets for programmatic weak supervision.

\subsection{Problem formulation}\label{sec:problem_formulation}

Given a dataset composed by $n$ instances with unknown labels \mbox{$y_1, y_2, \ldots, y_n \in \set{Y}~=\{1, 2, \ldots, T\}$}, programmatic weak supervision aims to use the outputs (guesses) of $m$ \acp{LF} to obtain probabilistic predictions for the label of each instance. The array composed by the guesses of the $m$ \acp{LF} at an instance is denoted by \mbox{$\Lambda = [L^{1}, L^{2}, \ldots, L^{m}]$}, where $L^j$ denotes the guess of the $j$-th \ac{LF}. Each array $\Lambda$ is composed by the outputs of multiple \acp{LF} that can provide guesses with assorted types such as labels ($L^{j} \in \set{Y}$), abstentions ($L^{j}=\mbox{`?'}$), or probabilities \mbox{($L^{j}=\hat{p}(1),\hat{p}(2),\ldots,\hat{p}(T)$)} (see Fig.~\ref{fig:intro}). In addition, for a specific instance $i\in\{1,2,\ldots,n\}$, we denote by \mbox{$\Lambda_i=[L_i^{1}, L_i^{2}, \ldots, L_i^{m}]$} the array given by the guesses for instance $i$.

Programmatic weak supervision aims to use the LFs' guesses $\Lambda_i$ to obtain probabilistic predictions $\up{h}_i$ for the label $y$ of each instance $i$. The probabilistic prediction for each instance $i\in\{1,2,\ldots,n\}$ is denoted by $\up{h}_{i}$ and given by $\up{h}_{i}~=~[\mathrm{h}_i(1), \mathrm{h}_i(2), \ldots, \mathrm{h}_i(T)]^\top\in\Delta(\set{Y})$, with $\Delta(\set{Y})$ the set of probability distributions on~$\set{Y}$.  The reliability of each probabilistic prediction $\up{h}_{i}$ can be roughly assessed by the \mbox{0-1 loss} $$\ell(\up{h}_{i},y_i)=\mathbb{I}\{\arg\max_y \up{h}_i(y)\neq y_i\}$$ or by a score tailored for probabilities, such as the log-loss $$\ell(\up{h}_{i},y_i)=-\log(\up{h}_{i}(y_i))$$ or the Brier-loss \cite{GneRaf:07} $$\ell(\up{h}_{i},y_i)=(1-\up{h}_{i}(y_i))^2.$$ Likewise, the expected loss at the $i$-th instance with respect to probability distribution $\up{p}_i\in\Delta(\set{Y})$ is given by $$\mathbb{E}_{\up{p}_i}\ell(\up{h}_i,y)=\sum_{y\in\set{Y}}\ell(\up{h}_i,y)\up{p}_i(y).$$

The reliability of probabilistic predictions can be also assessed by their similarity to the actual label probabilities for any group of instances. Specifically, let $\set{I}\subseteq \{1, 2, \ldots, n\}$ be a group of instances and $y \in \set{Y}$ be a label, the actual probability (proportion) of label $y$ for instances in $\set{I}$ is given by $$\up{p}^*_\set{I}(y) = \frac{1}{|\set{I}|} \sum_{i \in \set{I}} \mathbb{I}\{y_i = y\}$$ while the predicted probability of label $y$ for instances in $\set{I}$ is given by $$\up{h}_\set{I}(y) = \frac{1}{|\set{I}|} \sum_{i \in \set{I}} \up{h}_i(y).$$ Ideally, we would like that $\up{h}_\set{I}(y)\approx\up{p}^*_\set{I}(y)$ for any group $\set{I}$ and label $y$, and a significant difference between $\up{h}_\set{I}(y)$ and $\up{p}^*_\set{I}(y)$ implies that the probabilistic predictions for label $y$ are unreliable in group $\set{I}$. 

For each group $\set{I}\subseteq\{1,2,\ldots,n\}$ and label $y\in\set{Y}$, the actual probability $\up{p}^*_\set{I}(y)$ takes an unknown value in $[0,1]$. The methods presented in the following use the information provided by the \acp{LF} to determine a confidence interval for such probability, denoted by $[\,\underline{\up{p}}_\set{I}(y)\,,\, \overline{\up{p}}_\set{I}(y)\,]$. Notice that the methods that obtain confidence intervals in binary classification~\cite{barber2020distribution, gupta2020distribution} also provide intervals corresponding to groups of instances.

\subsection{Uncertainty sets that encapsulate the information of general LFs}\label{sec:general}

The proposed methodology for programmatic weak supervision is based on encapsulating the information provided by general \acp{LF} into an uncertainty set of distributions. 

Uncertainty sets are given by constraints on expectations of a feature mapping $\Phi(\Lambda, y) \in \mathbb{R}^d$ as 
 \begin{align}
   \label{eq:uncertainty set}
 \set{U} =\Big{\{} & \up{p}_1, \up{p}_2, \ldots, \up{p}_n\in\Delta(\set{Y}):\\\
 & \Big{|}\frac{1}{n}\sum_{i = 1}^n \mathbb{E}_{\up{p}_i} \big\{\Phi(\Lambda_{i}, y)\big\} - \widehat{\B{\tau}}\Big{|} \preceq \B{\lambda}\Big{\}}\nonumber
 \end{align}
where $\preceq$ and $| \cdot |$ denote component-wise inequalities and absolute values, vector $\widehat{\B{\tau}} \in \mathbb{R}^d$ denotes an estimate of the feature mapping's expectation $\B{\tau}=\frac{1}{n}\sum_{i=1}^n\Phi(\Lambda_i,y_i)$, and vector $\B{\lambda} \succeq \V{0}$ assesses the error in the expectation estimate. Similar uncertainty sets of distributions have been recently used to develop minimax approaches for supervised classification~\cite{fathony2016adversarial, mazuelas2020minimax, alvarez2023minimax, cortes2015structural, duchi2019variance}. 

The feature mapping $\Phi(\Lambda, y)$ allows us to encode multiple characteristics related to the \acp{LF}. For instance, if \mbox{$L^{1}, L^{2}, L^{3}\in\set{Y}$} are the outputs provided by 3 \acp{LF} that guess labels, their fit with labels can be encoded by the feature mapping 
\begin{align}
\label{eq:feature_freund}
\Phi(\Lambda, y) = \big[& \mathbb{I}\{L^{1} \neq y\}, \mathbb{I}\{L^{2} \neq y\}, \mathbb{I}\{L^{3} \neq y\}\big]^\top
\end{align}
for which the expectation $\B{\tau}$ corresponds to error probabilities for the different \acp{LF}. On the other hand, if $L^{1}\in\set{Y}$, \mbox{$L^{2}\in\set{Y}\cup\{\mbox{`?'}\}$}, and $L^{3}\in\Delta(\set{Y})$ are the outputs provided by 3 \acp{LF} that guess labels or label probabilities and may abstain at certain instances, their fit with labels and disagreements can be encoded by the feature mapping
\begin{align*}
\Phi(\Lambda, y) = \big[& \mathbb{I}\{L^1 \neq y\}, \mathbb{I}\{L^{2} \neq y\}, \mathbb{I}\{L^{2} \neq \mbox{`?'}\},\\
& (1-L^{3}(y))^2,\mathbb{I}\{L^{1} \neq L^{2}\},1-L^{3}(L^{1})\big]^\top
\end{align*}
for which the expectation $\B{\tau}$ corresponds to error and abstention probabilities together with expected scores and disagreements. In practice, the feature mapping used can be tailored to the \acp{LF} and information available, and 
Section~\ref{sec:uncertainty_sets} below discusses multiple choices for feature mappings components and corresponding expectation estimates.

In general, the uncertainty set $\set{U}$ in \eqref{eq:uncertainty set} contains all the probability distributions for labels that are consistent with the \acp{LF}' observed behavior, as measured by the feature mapping. The next sections show how such uncertainty sets can be used to obtain confidence intervals and probabilistic predictions.

\section{Confidence intervals for label probabilities using uncertainty sets}\label{sec:confidence_predictions}

This section shows how to obtain confidence intervals for the actual label probabilities from the information encapsulated in the uncertainty set.

Given an uncertainty set $\set{U}$ as in~(1), for any group of instances $\set{I} \subseteq \{1, 2, \ldots, n\}$ and label \mbox{$y \in \set{Y}$}, a confidence interval $[\,\underline{\up{p}}_\set{I}(y)\,,\, \overline{\up{p}}_\set{I}(y)\,]$ for the actual label probability $\up{p}^*_{\set{I}}(y)$  can be obtained using the optimum values
\begin{align}
\label{eq:confidence_interval_primal}
 \overline{\up{p}}_\set{I}(y) & = \underset{(\up{p}_{1}, \up{p}_{2}, \ldots, \up{p}_{n}) \in \set{U}}{\max} \; \frac{1}{|\set{I}|}\sum_{i \in \set{I}}\up{p}_i(y)\\% , \; \;  \; 
 \label{eq:confidence_interval_primal_2}
\underline{\up{p}}_\set{I}(y) & = \underset{(\up{p}_{1}, \up{p}_{2}, \ldots, \up{p}_{n}) \in \set{U}}{\min} \; \frac{1}{|\set{I}|}\sum_{i \in \set{I}}\up{p}_i(y).
 \end{align}
The upper  $\overline{\up{p}}_\set{I}(y)$ and the lower $\underline{\up{p}}_\set{I}(y)$ bounds represent, respectively, the maximum and minimum possible values allowed by $\set{U}$ for the average probability of label $y$ for instances in~$\set{I}$. The following result shows that such confidence interval can be determined by solving the optimization problems
\begin{align}
\label{eq:confidence_interval_upper} 
 \min_{\B{\mu}} \; \hspace{0.cm} &  - \widehat{\B{\tau}}^\top \B{\mu} +\B{\lambda}^\top |\B{\mu}|\\
 & + \frac{1}{n} \sum_{i = 1}^n \max_{\tilde{y} \in \set{Y}}\Big\{\Phi(\Lambda_i, \tilde{y})^\top \B{\mu} + \frac{n}{|\set{I}|}\mathbb{I}\{i\in\set{I}, \tilde{y} = y\}\Big\}\nonumber\\
\label{eq:confidence_interval_lower}
%\underline{\up{p}}_\set{I}(y) 
\max_{\B{\mu}} \;  &  - \widehat{\B{\tau}}^\top \B{\mu}-\B{\lambda}^\top |\B{\mu}|\\
& + \frac{1}{n} \sum_{i = 1}^n \min_{\tilde{y} \in \set{Y}}\,\Big\{\Phi(\Lambda_i, \tilde{y})^\top \B{\mu} + \frac{n}{|\set{I}|}\mathbb{I}\{i\in\set{I}, \tilde{y} = y\}\Big\}\nonumber
 \end{align}
{that correspond to the Lagrange dual of optimization problems in~\eqref{eq:confidence_interval_primal} and~\eqref{eq:confidence_interval_primal_2}.}

\begin{theorem}\label{th:confidence_interval} 
Let $\set{U}$ be a non-empty uncertainty set as in~\eqref{eq:uncertainty set} given by $\widehat{\B{\tau}}$ and $\B{\lambda}$, and let $\B{\tau}$ be the exact expectation of the feature mapping. % If $\overline{\up{p}}_\set{I}(y)$ and $\underline{\up{p}}_\set{I}(y)$ are given by~\eqref{eq:confidence_interval_primal}, then
For any group of instances $\set{I}$ and a label $y$, we have that $\overline{\up{p}}_\set{I}(y)$ and $\underline{\up{p}}_\set{I}(y)$ given by~\eqref{eq:confidence_interval_primal} and~\eqref{eq:confidence_interval_primal_2} are the minimum and the maximum values of the optimization problems~\eqref{eq:confidence_interval_upper} and~\eqref{eq:confidence_interval_lower}, respectively. In addition, if $\up{p}_\set{I}^*(y)$ is the actual label probability, we have that
\begin{align}
\label{eq:p1}
\up{p}_\set{I}^*(y) \in \big[\,\underline{\up{p}}_\set{I}(y)  - \underline{\varepsilon}\,,\, \overline{\up{p}}_\set{I}(y) + \overline{\varepsilon}\,\big]% \; \; 
\end{align}
\text{for \;$\overline{\varepsilon} =  (|\B{\tau} - \widehat{\B{\tau}}| - \B{\lambda})^\top |\overline{\B{\mu}}|$, \; $\underline{\varepsilon} = (|\B{\tau} - \widehat{\B{\tau}}| - \B{\lambda})^\top |\underline{\B{\mu}}|$} % , \; \; % \\
with $\overline{\B{\mu}}$ and $\underline{\B{\mu}}$ solutions to~\eqref{eq:confidence_interval_upper} and~\eqref{eq:confidence_interval_lower}, respectively. 
In particular, if $\B{\lambda}$ provides confidence intervals for the expectation with coverage probability $1 - \delta$, i.e., \mbox{$\mathbb{P}\{|\B{\tau} - \widehat{\B{\tau}}| \preceq \B{\lambda}\} \geq 1 - \delta$} we have that 
\begin{equation}
\label{eq:actual_prob_interval}
{\up{p}}_\set{I}^*(y) \in \big[\,\underline{\up{p}}_\set{I}(y)\,,\, \overline{\up{p}}_\set{I}(y)\,\big]
\end{equation}
with probability at least $1 - \delta$ simultaneously for all \mbox{$\set{I} \subseteq \{1,2, \ldots, n\}$} and $y\in\set{Y}$.
% \vspace{-0.2cm}
\begin{proof}
See Appendix~\ref{app:th_confidence_interval} in the supplementary materials.
\end{proof}
% \vspace{-0.3cm}
\end{theorem}

The theorem above shows that each confidence interval is given by the optimum values of the optimization problems~\eqref{eq:confidence_interval_upper} and~\eqref{eq:confidence_interval_lower}. These optimization problems can be efficiently addressed in practice since they are equivalent to linear optimization problems (defining new variables for the terms in the inner max and min in \eqref{eq:confidence_interval_upper} and \eqref{eq:confidence_interval_lower}). Furthermore, optimization problems \eqref{eq:confidence_interval_upper} and \eqref{eq:confidence_interval_lower} are also amenable for stochastic subgradient methods for convex optimization. In particular, the \acp{LF}' outputs corresponding to each instance $i\in\{1, 2, \ldots, n\}$ can provide subgradients for each term in the sums in \eqref{eq:confidence_interval_upper} and \eqref{eq:confidence_interval_lower} without the need to know the corresponding actual label. In addition, such optimization problems often have sparse solutions since the term $\B{\lambda}^\top|\B{\mu}|$ imposes an L1-regularization that penalizes feature components with high errors in expectation estimates.

 %\textcolor{blue}{Theorem~\ref{th:confidence_interval} shows that the validity of the presented confidence intervals is given by the reliability of the assessment for the error in mean estimates $(|\boldsymbol{\tau} - \widehat{\boldsymbol{\tau}}|)$. Large values of $\boldsymbol{\lambda}$ shrink the uncertainty set and lead to reduced intervals, but they do not guarantee that the actual probabilities are in the uncertainty set. On the other hand, small values of $\boldsymbol{\lambda}$ increase the likelihood that the uncertainty set contains the actual probability, but they result in wider confidence intervals. }
 
The optimal values $\overline{\up{p}}_\set{I}(y)$ and $\underline{\up{p}}_\set{I}(y)$ directly provide a valid confidence interval for label probabilities if the error in the expectation estimate is not underestimated,
i.e., \mbox{$\B{\lambda}\succeq|\B{\tau} - \widehat{\B{\tau}}|$}. In other cases, $\overline{\up{p}}_\set{I}(y)$ and $\underline{\up{p}}_\set{I}(y)$ still provide approximate confidence intervals as long as the underestimation \mbox{$|\B{\tau} - \widehat{\B{\tau}}| - \B{\lambda}$} is not substantial. Notice that the validity of the confidence intervals presented is independent of the instances' group and label considered, and only depends on the assessment for the error in expectation estimates. 

% \textcolor{blue}{The optimal values $\overline{\up{p}}_\set{I}(y)$ and $\underline{\up{p}}_\set{I}(y)$ directly provide a valid confidence interval for label probabilities if the error in the expectation estimate is not underestimated, i.e., \mbox{$\B{\lambda}\succeq|\B{\tau} - \widehat{\B{\tau}}|$}. Small values of $\boldsymbol{\lambda}$ increase the likelihood that the uncertainty set contains the actual probability, but they result in wider confidence intervals. In other cases, $\overline{\up{p}}_\set{I}(y)$ and $\underline{\up{p}}_\set{I}(y)$ still provide approximate confidence intervals as long as the underestimation \mbox{$|\B{\tau} - \widehat{\B{\tau}}| - \B{\lambda}$} is not substantial. Large values of~$\boldsymbol{\lambda}$ shrink the uncertainty set and lead to reduced intervals, but they do not guarantee that the actual probabilities are in the uncertainty set. Notice that the validity of the confidence intervals presented is independent of the instances' group and label considered, and only depends on the assessment for the error in expectation estimates. }

Theorem~\ref{th:confidence_interval} shows that the best value for $\B{\lambda}$ is the smallest value that ensures that the confidence intervals are reliable. This would occur by setting \mbox{$\B{\lambda} = |\B{\tau} - \widehat{\B{\tau}}|$}. If $\B{\lambda}$ is underestimated i.e.,  \mbox{$\B{\lambda}\preceq|\B{\tau} - \widehat{\B{\tau}}|$}, the error in the confidence intervals is at most \mbox{$\||\B{\tau} - \widehat{\B{\tau}}| - \B{\lambda}\| \max(\|\overline{\B{\mu}}\|, \|\underline{\B{\mu}}\|)$} as shown in equation~\eqref{eq:p1}. In practice, the value of $\B{\lambda}$ can be given by the an estimate of the error ($|\B{\tau} - \widehat{\B{\tau}}|$) such as the standard error.

% Reduced uncertainty sets given by more accurate expectation estimates or by additional feature components result in reduced feasible sets in~\eqref{eq:confidence_interval_primal} and~\eqref{eq:confidence_interval_primal_2} that lead to tighter confidence intervals.
The main factors that determine the width of confidence intervals are  the number of instances in the group and the size of the uncertainty set. More accurate expectation estimates,  additional feature components, and small values of $\B{\lambda}$ result in a reduced uncertainty set. In particular, adding feature components can shrink the uncertainty set if the new constrains added are informative (non-redundant features). Such uncertainty sets result in reduced feasible sets in~\eqref{eq:confidence_interval_primal} and~\eqref{eq:confidence_interval_primal_2} that lead to tighter confidence intervals. In addition, for sizable groups of instances $\set{I}$, the constraints in~\eqref{eq:confidence_interval_primal} and~\eqref{eq:confidence_interval_primal_2} more strongly affect the objective values leading to tighter confidence intervals. 

\subsection{Practical benefits of the confidence intervals presented} The intervals  $[\,\underline{\up{p}}_\set{I}(y)\,,\, \overline{\up{p}}_\set{I}(y)\,]$ can serve to assess the reliability of probabilistic predictions and the adequacy of \acp{LF} for different labels and groups of instances.

Confidence intervals for groups of instances with similar probabilistic predictions can be used to identify miscalibrated, overconfident, and underconfident predictions. For example, instances can be grouped for each label $y \in \set{Y}$ by using the probabilistic predictions as
\begin{equation}
    \label{eq:groups}
\set{I}_{p, {y}} = \Big{\{}i \in \{1, 2, \ldots, n\} :\up{h}_i(y) \geq p\Big{\}}
\end{equation}
\text{with $p \in [0, 1]$.} 
This type of groups can serve to assess the reliability of probabilistic predictions. For instance, a confidence interval of $[\,0.5\,,\, 0.7\,]$, for a group $\set{I}_{p, y}$ as in \eqref{eq:groups} with $p = 0.9$ and $y=1$, would indicate that probabilistic predictions larger than $0.9$ are highly unreliable for label $1$, and that the threshold~$0.9$ is inadequate to regard a label $1$ as correct.

Confidence intervals of groups of instances with similar \acp{LF}' outputs can be used to identify instances for which additional information is needed to reliable predict labels. For example, instances can be grouped by using the the proportion of \acp{LF} that predict the label $y \in \set{Y}$ as
\begin{equation}
    \label{eq:groups_LFs}
    \set{I}_{r, y} = \Big{\{}i \in \{1, 2, \ldots, n\} \colon \frac{1}{m}\sum_{j = 1}^m \mathbb{I}\big\{L_i^{j} = y\} \geq r\Big{\}}% \; 
\end{equation}
\text{with $r \in [0, 1]$.} 
% with $L_i^{(j)}$ the output of the $j$-th \ac{LF} on the $i$-th instance and $r \in [0, 1]$.
This type of groups can serve to assess the adequacy of \acp{LF}.
For instance, a confidence interval of~$[\,0.5\,,\, 0.95\,]$, for a group $\set{I}_{r, y}$ as in \eqref{eq:groups_LFs} with $r = 0.9$ and $y = 1$, would indicate that the \acp{LF} used are inadequate to accurately predict label $1$, prompting the need to obtain additional \acp{LF} or improve their prediction quality for label $1$.

The proposed confidence intervals have simultaneous coverage probability for all groups and labels since the validity of all intervals only depends on the assessment of errors in expectations estimates (\mbox{$|\B{\tau}-\widehat{\B{\tau}}|-\B{\lambda}$}). This capability is important in practice since it enables the usage of general groups of instances, even groups chosen using confidence intervals of other groups.    The process of identifying informative groups of instances can be iterative, focusing on the examination of overlapping groups and subgroups. For instance, after obtaining a confidence interval for a group $\set{I}$, we may want to examine the confidence intervals for subgroups $\set{I}' \subset \set{I}$ or for groups similar to $\set{I}$.  These procedures can help identify specific issues, such as challenging instances for labeling and the need for specific \acp{LF} for certain instances.

\subsection{Optimality of confidence intervals} Theorem~\ref{th:confidence_interval} shows that the validity of the presented confidence intervals is given by the assessment of error estimates (\mbox{$|\B{\tau}-\widehat{\B{\tau}}|-\B{\lambda}$}). The following provides bounds for the proposed confidence intervals in terms of the optimal confidence intervals $[\,\underline{\up{p}}_\set{I}^*(y)\,,\, \overline{\up{p}}_\set{I}^*(y)\,]$ given by
\begin{align}
 \label{eq:true_confidence_interval_upper}
 \overline{\up{p}}_\set{I}^*(y) & = \underset{(\up{p}_1,\up{p}_2,\ldots,\up{p}_n)\in\set{U}^*}{\max}\frac{1}{|\set{I}|}\sum_{i \in \set{I}} \up{p}_i(y)\\% , \; \; 
  \label{eq:true_confidence_interval_upper_2}
\underline{\up{p}}_\set{I}^*(y) & = \underset{(\up{p}_1,\up{p}_2,\ldots,\up{p}_n)\in\set{U}^*}{\min}\frac{1}{|\set{I}|}\sum_{i \in \set{I}} \up{p}_i(y)
 \end{align}
where $\set{U}^*$ is the uncertainty set
\begin{equation}
    \label{eq:smallest_uncertainty_set}
\set{U}^* =\Big{\{} \up{p}_1, \up{p}_2, \ldots, \up{p}_n\in\Delta(\set{Y}):\frac{1}{n}\sum_{i = 1}^n \mathbb{E}_{\up{p}_i} \{\Phi(\Lambda_{i}, y)\} = {\B{\tau}}\Big{\}}
\end{equation}
corresponding to an exact estimation of the expectation~$\B{\tau}$. For each group $\set{I}$ and label $y$, the confidence interval $[\,\underline{\up{p}}_\set{I}^*(y)\,,\, \overline{\up{p}}_\set{I}^*(y)\,]$ is optimal because it is contained in any confidence interval $[\,\underline{\up{p}}_\set{I}(y)-\underline{\varepsilon}\,,\, \overline{\up{p}}_\set{I}(y)+\overline{\varepsilon}\,]$ given by \eqref{eq:p1}. The next result provides bounds for the suboptimality of the confidence intervals presented.

\begin{theorem} \label{th:teorema2}
Let $\set{U}$ be a non-empty uncertainty set as in~\eqref{eq:uncertainty set} given by $\widehat{\B{\tau}}$ and $\B{\lambda}$, and $\B{\tau}$ be the exact expectation of the feature mapping. If $\overline{\up{p}}_\set{I}(y)$ and $\underline{\up{p}}_\set{I}(y)$ are given by~\eqref{eq:confidence_interval_primal} and~\eqref{eq:confidence_interval_primal_2} and $\overline{\up{p}}_\set{I}^*(y)$ and $\underline{\up{p}}_\set{I}^*(y)$ are given by~\eqref{eq:true_confidence_interval_upper} and~\eqref{eq:true_confidence_interval_upper_2} for a group of instances $\set{I} \subseteq  \{1, 2, \ldots, n\}$ and a label $y\in \set{Y}$. Then, we have that 
\begin{align}
\label{eq:optimal_interval}
    \! |\overline{\up{p}}_\set{I}(y) - \overline{\up{p}}_\set{I}^*(y)| & \leq \max(\|\B{\overline{\mu}}^* \|_1, \|\B{\overline{\mu}} \|_1) \||\B{\tau}  - \widehat{\B{\tau}}| + \B{\lambda}\|_\infty\\
    % |\overline{\up{p}}_\set{I}(y) - \overline{\up{p}}_\set{I}^*(y)| & \leq \max \; ( (|\B{\tau}  - \widehat{\B{\tau}}| + \B{\lambda})^\top |\B{\overline{\mu}}^*|,  \; (|\B{\tau}  - \widehat{\B{\tau}}| + \B{\lambda})^\top|\B{\overline{\mu}}|)\\
    \label{eq:optimal_interval2}
  \!   |\underline{\up{p}}_\set{I}(y) - \underline{\up{p}}_\set{I}^*(y)| & \leq \max(\|\B{\underline{\mu}}^*\|_1, \|\B{\underline{\mu}}\|_1) \||\B{\tau}  - \widehat{\B{\tau}}| + \B{\lambda}\|_\infty
   % |\underline{\up{p}}_\set{I}(y) - \underline{\up{p}}_\set{I}^*(y)| & \geq \min \; ((|\B{\tau}  - \widehat{\B{\tau}}| - \B{\lambda})^\top|\B{\underline{\mu}}^*|, \; (|\B{\tau}  - \widehat{\B{\tau}}| - \B{\lambda})^\top |\B{\underline{\mu}}|) 
\end{align}
where $\B{\overline{\mu}}^*$ and $\B{\underline{\mu}}^*$ are solutions to~\eqref{eq:confidence_interval_upper} and \eqref{eq:confidence_interval_lower}, respectively, taking $\widehat{\B{\tau}} = \B{\tau}$ and $\B{\lambda} = \V{0}$.
% \vspace{-0.2cm}
\begin{proof}
For the bound in~\eqref{eq:optimal_interval}, we have that 
\begin{align}
\overline{\up{p}}_\set{I}(y) = & 
 \min_{\B{\mu}} \; - \widehat{\B{\tau}}^\top \B{\mu}  + \B{\lambda}^\top |\B{\mu}|\\
 & + \frac{1}{n} \sum_{i = 1}^n \max_{\tilde{y} \in \set{Y}}\big{\{}\Phi(L_i, \tilde{y})^\top \B{\mu} + \frac{n}{|\set{I}|}\mathbb{I}\{i\in\set{I}, \tilde{y} = y\}\big{\}}\nonumber\\
 \leq & - \widehat{\B{\tau}}^\top \overline{\B{\mu}}^*+ \B{\lambda}^\top |\overline{\B{\mu}}^*|\\
 & + \frac{1}{n} \sum_{i = 1}^n \max_{\tilde{y} \in \set{Y}}\big{\{}\Phi(L_i, \tilde{y})^\top \overline{\B{\mu}}^* + \frac{n}{|\set{I}|}\mathbb{I}\{i\in\set{I}, \tilde{y} = y\}\big{\}}\nonumber
 \end{align}
with $\overline{\B{\mu}}^*$ solution to the optimization problem~\eqref{eq:confidence_interval_upper} taking $\widehat{\B{\tau}} = \B{\tau}$ and $\B{\lambda} = \V{0}$. Then, adding and subtracting $\B{\tau}^\top \overline{\B{\mu}}^*$, we have that 
\begin{align}
%\label{eq:overlinep}
\overline{\up{p}}_\set{I}(y) & \leq \overline{\up{p}}^*_\set{I}(y)  + ({\B{\tau}}- \widehat{\B{\tau}})^\top \overline{\B{\mu}}^* + \B{\lambda}^\top |\overline{\B{\mu}}^*|
 \end{align}
since $\overline{\up{p}}^*_\set{I}(y)$ is the minimum value of~\eqref{eq:confidence_interval_upper} taking $\widehat{\B{\tau}} = \B{\tau}$ and~$\B{\lambda} = \V{0}$. 
Hence, we have that
\begin{align}
\label{eq:overlinep}
\overline{\up{p}}_\set{I}(y)-\overline{\up{p}}^*_\set{I}(y)  \leq   \||\B{\tau}- \widehat{\B{\tau}}|+\B{\lambda}\|_\infty \|\overline{\B{\mu}}^*\|_1. 
 \end{align}
In addition, we have that 
\begin{align*}
\overline{\up{p}}^*_\set{I}(y) = & \min_{\B{\mu}} \; - {\B{\tau}}^\top \B{\mu}\\
& + \frac{1}{n} \sum_{i = 1}^n \max_{\tilde{y} \in \set{Y}}\big{\{}\Phi(\Lambda_i, \tilde{y})^\top \B{\mu} + \frac{n}{|\set{I}|}\mathbb{I}\{i\in\set{I}, \tilde{y} = y\}\big{\}} \\
& \leq - {\B{\tau}}^\top \overline{\B{\mu}}\\
&+ \frac{1}{n} \sum_{i = 1}^n \max_{\tilde{y} \in \set{Y}}\big{\{}\Phi(\Lambda_i, \tilde{y})^\top \overline{\B{\mu}} + \frac{n}{|\set{I}|}\mathbb{I}\{i\in\set{I}, \tilde{y} = y\}\big{\}}
 \end{align*}
with $\overline{\B{\mu}}$ solution to the optimization problem~\eqref{eq:confidence_interval_upper}. Then, adding and subtracting $\widehat{\B{\tau}}^\top \overline{\B{\mu}}$ and $\B{\lambda}^\top |\overline{\B{\mu}}|$, we have that 
\begin{align*}
\overline{\up{p}}^*_\set{I}(y) & \leq \overline{\up{p}}_\set{I}(y) + (\widehat{\B{\tau}} - {\B{\tau}})^\top \overline{\B{\mu}} - \B{\lambda}^\top |\overline{\B{\mu}}|
 \end{align*}
since $\overline{\up{p}}_\set{I}(y)$ is the minimum value of~\eqref{eq:confidence_interval_upper}. Hence, we have that 
\begin{align*}
\overline{\up{p}}^*_\set{I}(y)- \overline{\up{p}}_\set{I}(y)& \leq  \|| {\B{\tau}}-\widehat{\B{\tau}}|-\B{\lambda}\|_\infty\|\overline{\B{\mu}}\|_1\\
& \leq\|| {\B{\tau}}-\widehat{\B{\tau}}|+\B{\lambda}\|_\infty\|\overline{\B{\mu}}\|_1
%
%\overline{\up{p}}_\set{I}(y) & \geq \overline{\up{p}}^*_\set{I}(y) + ({\B{\tau}} - \widehat{\B{\tau}})^\top \overline{\B{\mu}} + \B{\lambda}^\top |\overline{\B{\mu}}|
 \end{align*}
 that together with bound in~\eqref{eq:overlinep} lead to~\eqref{eq:optimal_interval}. 

For the bound in~\eqref{eq:optimal_interval2}, we have that 
\begin{align*}
\underline{\up{p}}_\set{I}(y) = &
\max_{\B{\mu}} \; - \B{\lambda}^\top |\B{\mu}| - \widehat{\B{\tau}}^\top \B{\mu}\\
& + \frac{1}{n} \sum_{i = 1}^n \min_{\tilde{y} \in \set{Y}}\big{\{}\Phi(\Lambda_i, \tilde{y})^\top \B{\mu} + \frac{n}{|\set{I}|}\mathbb{I}\{i\in\set{I}, \tilde{y} = y\}\big{\}}\\
\geq & - \B{\lambda}^\top |\underline{\B{\mu}}^*| - \widehat{\B{\tau}}^\top \underline{\B{\mu}}^*\\
& + \frac{1}{n} \sum_{i = 1}^n \min_{\tilde{y} \in \set{Y}}\big{\{}\Phi(\Lambda_i, \tilde{y})^\top \underline{\B{\mu}}^* + \frac{n}{|\set{I}|}\mathbb{I}\{i\in\set{I}, \tilde{y} = y\}\big{\}}
\end{align*}
with $\underline{\B{\mu}}^*$ solution to the optimization problem~\eqref{eq:confidence_interval_lower} taking $\widehat{\B{\tau}} = \B{\tau}$ and $\B{\lambda} = \V{0}$. Then, adding and subtracting $\B{\tau}^\top \underline{\B{\mu}}^*$, we have that 
\begin{align}
%\label{eq:underlinep}
\underline{\up{p}}_\set{I}(y) & \geq\underline{\up{p}}^*_\set{I}(y) + (\B{\tau} - \widehat{\B{\tau}})^\top \underline{\B{\mu}}^*- \B{\lambda}^\top |\underline{\B{\mu}}^*| 
\end{align}
since $\underline{\up{p}}^*_\set{I}(y)$ is the maximum value of~\eqref{eq:confidence_interval_lower} taking $\widehat{\B{\tau}} = \B{\tau}$ and $\B{\lambda} = \V{0}$. Hence, we have that
\begin{align}
\label{eq:underlinep}
\underline{\up{p}}^*_\set{I}(y)-\underline{\up{p}}_\set{I}(y) & \leq  \||\B{\tau} - \widehat{\B{\tau}}|+\B{\lambda}\|_\infty\|\underline{\B{\mu}}^*\|_1.
\end{align}
In addition, we have that 
\begin{align*}
\underline{\up{p}}^*_\set{I}(y) \geq & \max_{\B{\mu}} \; - {\B{\tau}}^\top \B{\mu}\\
& + \frac{1}{n} \sum_{i = 1}^n \min_{\tilde{y} \in \set{Y}}\big{\{}\Phi(\Lambda_i, \tilde{y})^\top \B{\mu} + \frac{n}{|\set{I}|}\mathbb{I}\{i\in\set{I}, \tilde{y} = y\}\big{\}}\\
\geq & - {\B{\tau}}^\top \underline{\B{\mu}}\\
& + \frac{1}{n} \sum_{i = 1}^n \min_{\tilde{y} \in \set{Y}}\big{\{}\Phi(\Lambda_i, \tilde{y})^\top \underline{\B{\mu}} + \frac{n}{|\set{I}|}\mathbb{I}\{i\in\set{I}, \tilde{y} = y\}\big{\}}
\end{align*}
with $\underline{\B{\mu}}$ solution to the optimization problem~\eqref{eq:confidence_interval_lower}. Then, adding and subtracting $\widehat{\B{\tau}}^\top \underline{\B{\mu}}$ and $\B{\lambda}^\top |\underline{\B{\mu}}|$, we have that 
\begin{align}
\underline{\up{p}}^*_\set{I}(y) & \geq \underline{\up{p}}_\set{I}(y)+ (\widehat{\B{\tau}} - {\B{\tau}})^\top \underline{\B{\mu}} + \B{\lambda}^\top |\underline{\B{\mu}}| 
\end{align}
since $\underline{\up{p}}_\set{I}(y)$ is the maximum value of~\eqref{eq:confidence_interval_lower}.

Hence, we have that 
\begin{align*}
\underline{\up{p}}_\set{I}(y) -\underline{\up{p}}^*_\set{I}(y)&\leq   \||{\B{\tau}} - \widehat{\B{\tau}}|-\B{\lambda}\|_\infty\| \underline{\B{\mu}}\|_1\\
&\leq\||{\B{\tau}} - \widehat{\B{\tau}}|+\B{\lambda}\|_\infty\| \underline{\B{\mu}}\|_1
%- \B{\lambda}^\top |\underline{\B{\mu}}| 
\end{align*}
that together with bound in~\eqref{eq:underlinep} lead to~\eqref{eq:optimal_interval2}.
 \end{proof}
 % \vspace{-0.2cm}
\end{theorem}
The theorem above bounds the difference between the proposed and optimal confidence intervals for each group $\set{I}$ and label $y \in \set{Y}$. In particular, bounds \eqref{eq:optimal_interval} and \eqref{eq:optimal_interval2} show that the suboptimality of the confidence intervals proposed is of the order of the error in the expectation estimates.

The confidence intervals presented are agnostic to the method used to obtain probabilistic predictions, and can be used to assess the reliability of general 
predictions. The next section presents methods to obtain probabilistic predictions that are ensured to belong to such confidence intervals. 

\section{Minimax probabilistic predictions using uncertainty sets}\label{sec:prob_pred}
% \vspace{-0.2cm}
This section shows how the uncertainty sets can also be
used to obtain more reliable probabilistic predictions. Given the uncertainty set $\set{U}$ in~\eqref{eq:uncertainty set}, the proposed methodology obtains probabilistic predictions $(\up{h}_{1}, \up{h}_{2}, \ldots, \up{h}_n)$ by minimizing the worst-case expected log-loss over all instances. These probabilities are solutions to the minimax problem 
 % \vspace{-0.1cm}
 \begin{equation}
 \label{eq:minmax}
\underset{(\up{h}_{1}, \up{h}_{2}, \ldots, \up{h}_{n})}{\min} \, \, \underset{(\up{p}_{1}, \up{p}_{2}, \ldots, \up{p}_{n}) \in \set{U}}{\max} \; \frac{1}{n}\sum_{i = 1}^{n}\mathbb{E}_{\up{p}_i}\{\ell(\up{h}_{i}, y)\}
\end{equation}
where $\mathbb{E}_{\up{p}}\{\ell(\up{h}, y)\}$ denotes the expected log-loss of $\up{h}\in\Delta(\set{Y})$ with respect to $\up{p}\in\Delta(\set{Y})$, i.e., $$\mathbb{E}_{\up{p}}\{\ell(\up{h}, y)\}=\sum_{y\in\set{Y}}\ell(\up{h}, y)\up{p}(y)=-\sum_{y\in\set{Y}}\log(\up{h}(y))\up{p}(y).$$ The usage of log-loss is adequate not only for analytical convenience but also because the log-loss is a strictly proper score for probabilities \cite{gneiting2007strictly}. In particular, the usage of such loss ensures  that the  probabilistic predictions are always contained in the confidence intervals provided, as shown in Theorem~\ref{th:prob_estimated} below.

The following result shows that the minimax probabilities can be determined by solving the optimization problem
% \vspace{-0.1cm}
\begin{align}
\label{eq:optim_problem}
\min_{\B{\mu}} \; \; &  - \widehat{\B{\tau}}^\top \B{\mu}+\B{\lambda}^\top |\B{\mu}|\\
& + \frac{1}{n}\sum_{i = 1}^n \log\Big{(}\sum_{\tilde{y} \in \set{Y}} \exp \left\{\Phi(\Lambda_{i}, \tilde{y})^\top \B{\mu} \right\}\Big{)}.\nonumber
\end{align}

\begin{theorem}\label{th:estimated_probability}
Let $\set{U}$ be a non-empty uncertainty set as in~\eqref{eq:uncertainty set} given by $\widehat{\B{\tau}}$ and $\B{\lambda}$. If $\B{\mu}^*$ is a solution of the optimization problem~\eqref{eq:optim_problem} and probabilistic predictions are given by
\begin{equation}
    \label{eq:prob}
   \up{h}_i({y}) = \Big{(}\sum_{\tilde{y} \in \set{Y}} \exp\big\{ (\Phi(\Lambda_i, \tilde{y}) - \Phi(\Lambda_i, {y}))^\top\B{\mu}^*\big\}\Big{)}^{-1}\\%, \mbox{ for }i = 1, 2, \ldots, n.
\end{equation} 
for $i = 1, 2, \ldots, n$. Then, $(\up{h}_{1}, \up{h}_2, \ldots, \up{h}_n)$ form a solution of the minimax problem~\eqref{eq:minmax}.
In addition, the minimum value of~\eqref{eq:optim_problem} equals the minimax risk value of~\eqref{eq:minmax} that is denoted in the following as~$R(\set{U})$.
% \vspace{-0.3cm}
\begin{proof}
The result can be proven analogously of that in Theorem 5 in~\cite{mazuelas2022entropy}. Firstly, for $(\up{h}_1,\up{h}_2,\ldots,\up{h}_n)$ such that $\up{h}_i \in \Delta(\set{Y})$ for $i = 1, 2, \ldots, n$, we have that
\begin{align}
\label{eq:proof_optim_problem}
     & \hspace{-0cm}\max_{(\up{p}_1, \up{p}_2, \ldots, \up{p}_n) \in \set{U}} && \frac{1}{n}\sum_{i = 1}^n \mathbb{E}_{\up{p}_i}\{\ell(\up{h}_i, y)\}\\
     = & \max_{(\up{p}_1, \up{p}_2, \ldots, \up{p}_n) \in \set{U}} &&  \frac{1}{n}\B{\up{l}}^\top \B{\up{p}} - I_+(\B{\up{p}})\\
    & \hspace{0.8cm}\text{s. t.}&&  \sum_{y \in \set{Y}} \up{p}_i(y) = 1, \; i = 1, 2, \ldots, n\nonumber\\
    &&& \widehat{\B{\tau}} - \B{\lambda}\preceq\frac{1}{n}\B{\Phi}^\top\B{\up{p}} \preceq \widehat{\B{\tau}} + \B{\lambda}\nonumber
\end{align}
where \begin{align*}
    I_+(\B{\up{p}}) = \left\{
    \begin{array}{cc}
       0  & \text{ if } \B{\up{p}}\succeq \V{0} \\
        \infty & \text{ otherwise}
    \end{array} \right.
\end{align*}
and $\B{\up{p}}, \B{\up{l}}$, and $\B{\Phi}$ denote the vectors and matrix composed by concatenating for each $y \in \set{Y}$ the vectors and matrix with rows $\up{p}_i(y), - \log(\up{h}_i(y)),$ and $\Phi(\Lambda_i, y)^\top$ for $i = 1, 2, \ldots, n$. 

Optimization problem~\eqref{eq:proof_optim_problem} has Lagrange dual
\begin{align*}
    & \min_{\B{\mu}_1, \B{\mu}_2, \B{\nu}} && - (\widehat{\B{\tau}} - \B{\lambda})^\top \B{\mu}_1 + (\widehat{\B{\tau}} + \B{\lambda})^\top \B{\mu}_2 + \sum_{i = 1}^n \nu^{(i)}\\
    &&&+ f^*\Big(\B{\Phi}\frac{(\B{\mu}_1 - \B{\mu}_2)}{n} - {\B{\nu}}\Big)\\
    & \hspace{0.25cm} \text{s. t.} \hspace{-2cm}&& \B{\mu}_1, \B{\mu}_2 \succeq \V{0}
\end{align*}
where ${\B{\nu}}$ is the vector in $\mathbb{R}^{n |\set{Y}|}$ with component corresponding with the pairs ($\Lambda_i$, $y$) for \mbox{$i = 1, 2, \ldots, n$},  \mbox{$y \in \set{Y}$} given by $\nu^{(i)}$, and $f^*$ is the conjugate function of \mbox{$f(\B{\up{p}}) = -(1/n) \B{\up{l}}^\top \B{\up{p}} + I_+(\B{\up{p}})$} given by 
\begin{equation*}
 f^*(\V{w}) = \max_{\B{\up{p}}\succeq \V{0}} \; \V{w}^\top \B{\up{p}} + \frac{1}{n}\B{\up{l}}^\top \B{\up{p}} = \left\{\begin{array}{cc}
     0 & \text{ if } \V{w} \preceq - (1/n)\B{\up{l}} \\
     \infty & \text{ otherwise.}
 \end{array}   \right.
\end{equation*}
Therefore, the Lagrange dual above becomes
\begin{align*}
    & \min_{\B{\mu}_1, \B{\mu}_2, \B{\nu}} && - (\widehat{\B{\tau}} - \B{\lambda})^\top \B{\mu}_1 + (\widehat{\B{\tau}} + \B{\lambda})^\top \B{\mu}_2 +  \sum_{i = 1}^n \nu^{(i)}\\
    & \text{s. t.} && \B{\mu}_1, \B{\mu}_2 \succeq \V{0}\\
    &&&\frac{1}{n}\Phi(\Lambda_i, y)^\top(\B{\mu}_1 - \B{\mu}_2) - \nu^{(i)} \leq \frac{1}{n}\log(\up{h}_i(y)),\\
    &&&\forall \ y \in \set{Y}, i = 1, 2, \ldots, n.%\sum_{y \in \set{Y}}\exp\{\Phi(\Lambda_i, y)^\top\B{\mu} + \nu\}\leq 1,
\end{align*}
The solution of such optimization problem $\B{\mu}_1^*, \B{\mu}_2^*$ satisfies ${\mu_1^*}^{(i)}{\mu_2^{*}}^{(i)} = 0$ for any $i$ such that \mbox{$\lambda^{(i)} > 0$}. Then, \mbox{$\B{\lambda}^\top(\B{\mu}_1^* + \B{\mu}_2^*) = \B{\lambda}^\top|\B{\mu}_1^* - \B{\mu}_2^*|$} and taking $\B{\mu} =\B{\mu}_{1} - \B{\mu}_2$, the Lagrange dual above is equivalent to
\begin{align*}
    & \min_{\B{\mu}, \B{\nu}} && - \widehat{\B{\tau}}^\top \B{\mu} + \B{\lambda}^\top |\B{\mu}| + \sum_{i = 1}^n \nu^{(i)}\\
    & \text{s. t.}  &&\frac{1}{n}\Phi(\Lambda_i, y)^\top\B{\mu} - \nu^{(i)} \leq \frac{1}{n}\log(\up{h}_i(y)),\\
    &&&\forall \ y \in \set{Y}, i = 1, 2, \ldots, n%\sum_{y \in \set{Y}}\exp\{\Phi(\Lambda_i, y)^\top\B{\mu} + \nu\}\leq 1,
\end{align*}
that has the same value as $$\max_{(\up{p}_1, \up{p}_2, \ldots, \up{p}_n)\in\set{U}} \frac{1}{n}\sum_{i = 1}^n\mathbb{E}_{\up{p}_i}\{\ell(\up{h}_i, y)\}$$ since the constraints in~\eqref{eq:proof_optim_problem} are affine and $\set{U}$ is non-empty. Therefore, we have that 
\begin{align}
& \min_{(\up{h}_1, \up{h}_2, \ldots, \up{h}_n)}  \max_{(\up{p}_1, \up{p}_2, \ldots, \up{p}_n) \in \set{U}}  \frac{1}{n}\sum_{i = 1}^n \mathbb{E}_{\up{p}_i}\{\ell(\up{h}_i, y)\}  \nonumber\\
\label{eq:min_app2}
 &= \min_{(\up{h}_1, \up{h}_2, \ldots, \up{h}_n), \B{\mu}, \B{\nu}} - \widehat{\B{\tau}}^\top \B{\mu} + \B{\lambda}^\top|\B{\mu}| + \sum_{i = 1}^n \nu^{(i)}\\
    &\hspace{1.1cm}\text{ s. t. } \hspace{0.85cm} \frac{1}{n}\Phi(\Lambda_i, y)^\top\B{\mu} - \nu^{(i)} \leq \frac{1}{n}\log(\up{h}_i(y)), \nonumber\\
    &\hspace{2.8cm}\forall \ y \in \set{Y}, i = 1, 2, \ldots, n. \nonumber%\sum_{y \in \set{Y}}\exp\{\Phi(\Lambda_i, y)^\top\B{\mu} + \nu\}\leq 1.
\end{align}
The constraints in the above optimization problem satisfy
\begin{equation}
\label{eq:phi_lambdai}
\Phi(\Lambda_i, y)^\top \B{\mu} - n\nu^{(i)} \leq 
 \log (\up{h}_i(y)), \; \forall \ y \in \set{Y}, i = 1, 2, \ldots, n
 \end{equation}
 so that, for any $y\in \set{Y}$ and $i=1,2, \ldots, n$, we have that
 \begin{equation*}
\exp\big{\{}\Phi(\Lambda_i, y)^\top\B{\mu} - n\nu^{(i)}\big{\}} \leq 
 \up{h}_i(y)
  \end{equation*}
  hence
   \begin{equation*}
 \sum_{y \in \set{Y}} \exp\big{\{}\Phi(\Lambda_i, y)^\top\B{\mu}  - n\nu^{(i)}\big{\}} \leq 
 1, \; i = 1, 2, \ldots, n
 \end{equation*}
 that implies
    \begin{align*}
% \label{eq:constraints}
\nu^{(i)} \geq \frac{1}{n}\log\Big{(}\sum_{y \in \set{Y}}\exp\{\Phi(\Lambda_i, y)^\top\B{\mu}\}\Big{)}, \; i = 1, 2, \ldots, n.
\end{align*}
Then, for each $\B{\mu}$, 
taking 
\begin{align}
\label{eq:proof_estim_prob}
\up{h}_i(y) & = \exp\Big{\{}\Phi(\Lambda_i, y)^\top\B{\mu} - \log\Big{(}\sum_{\tilde{y} \in \set{Y}}\exp\{\Phi(\Lambda_i, \tilde{y})^\top\B{\mu}\}\Big{)}\Big{\}}\\
\nu^{(i)} & = \frac{1}{n}\log\Big{(}\sum_{y \in \set{Y}}\exp\{\Phi(\Lambda_i, y)^\top\B{\mu}\}\Big{)}\nonumber
\end{align}
we have that any probability distribution given by~\eqref{eq:proof_estim_prob} is solution of 
\begin{align*}
    \min_{(\up{h}_{1}, \up{h}_{2}, \ldots, \up{h}_{n}), \B{\nu}} \; \; \; & \sum_{i = 1}^n \nu^{(i)}\\
    &  \frac{1}{n}\Phi(\Lambda_i, y)^\top \B{\mu} -  \nu^{{(i)}}\leq \frac{1}{n}\log (\up{h}_i(y)),\\
    & \forall \ y\in \set{Y}, i = 1, 2, \ldots, n.
\end{align*}
that has optimum value
\begin{align*}
 \frac{1}{n}\sum_{1=1}^n\log\Big{(}\sum_{\tilde{y} \in \set{Y}}\exp\{\Phi(\Lambda_i, \tilde{y})^\top\B{\mu}\}\Big{)}
\end{align*}
by substituting $\up{h}_{i}$ and $\nu^{(i)}$. 
Then, the optimization problem in~\eqref{eq:optim_problem} is obtained by substituting the above optimization problem in~\eqref{eq:min_app2}.
\end{proof}
% \vspace{-0.4cm}
\end{theorem}

The theorem above shows that minimax probabilistic predictions are given by combinations of the components of the feature mapping $\Phi(\Lambda,y)$ that represents the \acp{LF}' guesses $\Lambda$ for different labels~$y$. In the following, we refer to the probabilistic predictions in~\eqref{eq:prob} as \acp{WMRC}. 
Existing minimax methods obtain probabilistic predictions corresponding to uncertainty sets given by feature mappings as in~\eqref{eq:feature_freund} that represent the error of \acp{LF} that guess labels \cite{balsubramani2016optimal,Steven:2024}, while Theorem~\ref{th:estimated_probability} extends those results for general feature mappings and \acp{LF}. In particular, as depicted in Section~\ref{sec:general} and detailed in Section~\ref{sec:uncertainty_sets}, the \acp{WMRC} given by \eqref{eq:prob} can provide reliable probabilistic predictions for \acp{LF} that provide guesses with assorted types, e.g., \acp{LF} that not only provide label guesses but can also abstain in certain instances or provide guesses for label probabilities.

The vector parameter $\B{\mu}^{*} \in \mathbb{R}^{d}$ determining \acp{WMRC} is obtained by solving the convex optimization problem in~\eqref{eq:optim_problem}, which can be solved efficiently similarly to those in \eqref{eq:confidence_interval_upper} and \eqref{eq:confidence_interval_lower} above.  In particular, optimization problem \eqref{eq:optim_problem} is convex and amenable for stochastic subgradient methods that only require to evaluate \acp{LF}' guesses in different instances. In addition, vector $\B{\mu}^{*}$ is often sparse due to the L1-regularization imposed by the term $\B{\lambda}^\top|\B{\mu}|$.

The following further extends the theoretical characterization of \acp{WMRC} by providing performance guarantees in terms of minimax risks and differences with actual label probabilities.

\begin{theorem}\label{th:prob_estimated}
Let $\set{U}$ be a non-empty uncertainty set given by $\widehat{\B{\tau}}$ and $\B{\lambda}$ as in~\eqref{eq:uncertainty set}, and $\B{\tau}$ be the exact expectation of the feature mapping. If $\up{h}_1, \up{h}_2, \ldots, \up{h}_n$ are the probabilistic predictions in~\eqref{eq:prob} determined by vector parameter $\B{\mu}^*$,  their actual \mbox{log-loss} satisfy
\begin{align}
\label{eq:bound_risk}
% &\frac{1}{n} \sum_{i = 1}^n R(\up{h}_i) \leq R(\set{U}) + \||\B{\tau}  - \widehat{\B{\tau}}|- \B{\lambda}\|_\infty \|\B{\mu}^*\|_1.
&\frac{1}{n} \sum_{i = 1}^n \ell(\up{h}_i, y_i) \leq R(\set{U}) + (|\B{\tau}  - \widehat{\B{\tau}}|- \B{\lambda})^\top |\B{\mu}^*|\,.
\end{align}
Furthermore, for any group $\set{I}\subseteq\{1,2,\ldots,n\}$ and label $y\in\set{Y}$, we have that \mbox{$\up{h}_{\set{I}}(y)=\frac{1}{|\set{I}|}\sum_{i\in\set{I}}\up{h}_i(y)$} is included in the confidence interval defined by \eqref{eq:confidence_interval_primal}, (i.e., \mbox{$\up{h}_\set{I}(y) \in [\,\underline{\up{p}}_\set{I}(y)\,,\, \overline{\up{p}}_\set{I}(y)\,]$}), and its difference with respect to the actual label probability satisfies
\begin{align}
\label{eq:h}
|\up{h}_\set{I}(y) - \up{p}_\set{I}^*(y)| \leq \max\big(& \overline{\up{p}}_\set{I}(y)-\up{h}_\set{I}(y) +\overline{\varepsilon},\\ 
&\up{h}_\set{I}(y)- \underline{\up{p}}_\set{I}(y) +\underline{\varepsilon} \big)\nonumber% \\
\end{align}
with $\overline{\varepsilon}=(|\B{\tau} - \widehat{\B{\tau}}| - \B{\lambda})^\top |\overline{\B{\mu}}|$ and $\underline{\varepsilon}=(|\B{\tau} - \widehat{\B{\tau}}| - \B{\lambda})^\top |\underline{\B{\mu}}|$, as in \eqref{eq:p1} of Theorem~\ref{th:confidence_interval} .

\begin{proof}
Let $\set{U}^*$ be the uncertainty set given by \eqref{eq:uncertainty set} for \mbox{$\widehat{\B{\tau}}=\B{\tau}=\frac{1}{n}\sum_{i = 1}^n \Phi(\Lambda_i, y_i)$} and $\B{\lambda}=\V{0}$. Such uncertainty is non-empy since it includes the distributions given by actual labels \mbox{$\mathbb{I}\{y=y_1\},\mathbb{I}\{y=y_2\},\ldots,\mathbb{I}\{y=y_n\}\in\Delta(\set{Y})$}. Due to such an inclusion, the actual log-losses satisfy
\begin{align}\label{step-opt}\frac{1}{n} \sum_{i = 1}^n \ell(\up{h}_i,y_i) \leq \underset{(\up{p}_1, \up{p}_2, \ldots, \up{p}_n) \in \mathcal{U}^*}{\max} \; \frac{1}{n}\sum_{i = 1}^n \mathbb{E}_{\up{p}_i}\ell (\up{h}_i, y)\, .\end{align}
The optimization problem in the right-hand-side of \eqref{step-opt} has a Lagrange dual given by
%\begin{align}\label{primal_aux}\underset{(\up{p}_1, \up{p}_2, \ldots, \up{p}_n) \in \mathcal{U}^*}{\max} \; \ell (\up{h}_i, \up{p}_i)\end{align}
%has Lagrange dual given by
\begin{align*}
% \begin{matrix}
&\underset{\B{\mu}, \B{\nu}}{\min} \hspace{0cm}&& - {{\B{\tau}}}^\top \B{\mu}  + \sum_{i = 1}^n\nu^{(i)}\\
&\text{\,s. t. } \hspace{0cm}&& \Phi(\Lambda_i, y)^\top \B{\mu} - n\nu^{(i)} \leq \log(\up{h}_i(y))\\
&&& \forall \; y \in \mathcal{Y}, i = 1, 2, \ldots, n\,.% \end{matrix}
\end{align*}

%includes the actual probabilities $\{\up{p}_1^*,\up{p}_2^*,\ldots,\up{p}_n^*\}$. 
 
%For the bound in~\eqref{eq:bound_risk}, we have that 
%$$\frac{1}{n} \sum_{i = 1}^n R(\up{h}_i) \leq \underset{(\up{p}_1, \up{p}_2, \ldots, \up{p}_n) \in \mathcal{U}^*}{\max} \; \frac{1}{n}\sum_{i = 1}^n \ell (\up{h}_i, \up{p}_i)$$
%since $R(\up{h}_i) =  \frac{1}{n}\sum_{i = 1}^n  \ell(\up{h}_i, \up{p}_i^*)$ and $\up{p}_i^* \in \mathcal{U}^*$ by definition of $\mathcal{U}^*$. 

The constraints in the above optimization problem satisfy
\begin{equation*}
\Phi(\Lambda_i, y)^\top \B{\mu} - n\nu^{(i)} \leq 
 \log (\up{h}_i(y)), \; \forall \ y \in \set{Y}, i = 1, 2, \ldots, n
 \end{equation*}
 that as done in~\eqref{eq:phi_lambdai} leads to
\begin{equation}
\label{eq:constraints2}
\nu^{(i)} \geq \frac{1}{n}\log\Big{(}\sum_{y \in \set{Y}}\exp\{\Phi(\Lambda_i, y)^\top\B{\mu}\}\Big{)}, \; i = 1, 2, \ldots, n\, .
\end{equation}
Then, since the constraints in \eqref{step-opt} are linear-affine and the uncertainty set $\set{U}^*$ is non-empty, strong duality is attained and we have
\begin{align*}
&\hspace{-0.2cm}\underset{(\up{p}_1, \up{p}_2, \ldots, \up{p}_n) \in \mathcal{U}^*}{\max}  \; \frac{1}{n}\sum_{i = 1}^n\mathbb{E}_{\up{p}_i}\{\ell (\up{h}_i, y)\}\\
&\hspace{0.6cm}= \underset{\B{\mu}}{\min} \; - {{\B{\tau}}}^{\top} \B{\mu} + \frac{1}{n} \sum_{i=1}^n \log\Big{(}\sum_{y \in \set{Y}}\exp\{\Phi(\Lambda_i, y)^\top\B{\mu}\}\Big{)}\\
&\hspace{0.6cm}\leq  - {{\B{\tau}}}^{\top} \B{\mu}^* + \frac{1}{n} \sum_{i=1}^n \log\Big{(}\sum_{y \in \set{Y}}\exp\{\Phi(\Lambda_i, y)^\top\B{\mu}^*\}\Big{)}.
\end{align*}
%that results in
%\begin{align*}
%\underset{(\up{p}_1, \up{p}_2, \ldots, \up{p}_n) \in \mathcal{U}^*}{\max} \; \frac{1}{n}\sum_{i = 1}^n\mathbb{E}_{\up{p}_i}\ell (\up{h}_i, y) & = \underset{\B{\mu}}{\min} \; - {{\B{\tau}}}^{\top} \B{\mu} + \frac{1}{n} \sum_{i=1}^n \log\Big{(}\sum_{y \in \set{Y}}\exp\{\Phi(\Lambda_i, y)^\top\B{\mu}\}\Big{)} \nonumber\\
%& \leq  - {{\B{\tau}}}^{\top} \B{\mu}^* + \frac{1}{n} \sum_{i=1}^n \log\Big{(}\sum_{y \in \set{Y}}\exp\{\Phi(\Lambda_i, y)^\top\B{\mu}^*\}\Big{)}. 
%\end{align*}
Then, we have 
\begin{align}
\frac{1}{n} \sum_{i = 1}^n &\ell(\up{h}_i,y_i) \nonumber\\
\leq & - {{\B{\tau}}}^{\top} \B{\mu}^* + \frac{1}{n} \sum_{i=1}^n \log\Big{(}\sum_{y \in \set{Y}}\exp\{\Phi(\Lambda_i, y)^\top\B{\mu}^*\}\Big{)} \nonumber\\
= & - {{\B{\tau}}}^{\top} \B{\mu}^* + \frac{1}{n} \sum_{i=1}^n \log\Big{(}\sum_{y \in \set{Y}}\exp\{\Phi(\Lambda_i, y)^\top\B{\mu}^*\}\Big{)}\\
& + {\widehat{\B{\tau}}}^\top \B{\mu}^* - {\widehat{\B{\tau}}}^{\top} \B{\mu}^* + \B{\lambda}^{\top} |\B{\mu}^*| - \B{\lambda}^{\top}|\B{\mu}^*|  \nonumber\\
\label{eq:20}
= & R(\mathcal{U}) - \B{\tau}^{\top} \B{\mu}^* + {\widehat{\B{\tau}}}^{\top} \B{\mu}^*  - \B{\lambda}^{\top}|\B{\mu}^*|
 \end{align}
 that leads to the bound in~\eqref{eq:bound_risk} using H\"{o}lder's inequality.
 
To prove that $\up{h}_\set{I}(y) \in [\underline{\up{p}}_\set{I}(y), \overline{\up{p}}_\set{I}(y)]$, we use that $(\up{h}_1, \up{h}_{2}, \ldots, \up{h}_{n})$ solutions of~\eqref{eq:minmax} are included in the uncertainty set $\set{U}$ because $(\up{h}_1, \up{h}_{2}, \ldots, \up{h}_{n})$ coincide with the worst-case distribution in $\set{U}$. Such result is obtained because the log-loss is strictly proper, and the min and the max in \eqref{eq:minmax} can be interchanged since the objective function is lower semi-continuous and $\set{U}$ is compact and convex in $\mathbb{R}^{n|\set{Y}|}$ \cite{simons1995minimax}. Then, for any $\up{h}_{i}$ and $y \in \set{Y}$, we have \begin{align*}
\min_{(\up{p}_{1}, \up{p}_{2}, \ldots, \up{p}_{n}) \in \set{U}} \frac{1}{|\set{I}|} \sum_{i \in \set{I}} \up{p}_i(y) & \leq \frac{1}{|\set{I}|}\sum_{i \in \set{I}} \up{h}_{i}(y)\\
& \leq \max_{(\up{p}_{1}, \up{p}_{2}, \ldots, \up{p}_{n})\in \set{U}} \frac{1}{|\set{I}|} \sum_{i \in \set{I}} \up{p}_i(y)
\end{align*}
that directly leads $\up{h}_\set{I}(y) \in [\underline{\up{p}}_\set{I}(y), \overline{\up{p}}_\set{I}(y)]$.

%since the optimization problems in \eqref{eq:confidence_interval_upper} and \eqref{eq:confidence_interval_lower} are the dual of the optimization problems in the inequalities above.

Hence,  the inequality in \eqref{eq:h} is obtained because \mbox{$\up{h}_\set{I}(y) \in [\underline{\up{p}}_\set{I}(y), \overline{\up{p}}_\set{I}(y)]$} together with the bound in \eqref{eq:p1} of Theorem~\ref{th:confidence_interval} imply
$$\up{p}_\set{I}^*(y) -\up{h}_\set{I}(y) \leq  \overline{\up{p}}_\set{I}(y)- \up{h}_\set{I}(y) + (|\widehat{\B{\tau}}-\B{\tau}| - \B{\lambda})^\top|\overline{\B{\mu}}|$$
$$\up{h}_\set{I}(y)-\up{p}_\set{I}^*(y)  \leq  \up{h}_\set{I}(y)-\underline{\up{p}}_\set{I}(y)  + (|\widehat{\B{\tau}}-\B{\tau}| - \B{\lambda})^\top|\underline{\B{\mu}}|$$
that directly leads to the bound in the theorem.
    \end{proof}
\end{theorem}

The theorem above provides bounds for the actual log-loss of \acp{WMRC} in terms of the minimax risk optimized in~\eqref{eq:optim_problem}, and for the differences between \acp{WMRC} and actual label probabilities of any group of instances. Similarly to Theorem~\ref{th:confidence_interval}, minimax risks and confidence intervals provide valid bounds for the actual log-loss and differences with actual label probabilities in cases where the error in the expectation estimate satisfies $\B{\lambda} \succeq|\B{\tau} - \widehat{\B{\tau}}|$. In other cases, minimax risks and confidence intervals still provide approximate bounds as long as the underestimation $|\B{\tau} - \widehat{\B{\tau}}| - \B{\lambda}$ is not substantial.

The proposed \ac{WMRC} can provide more reliable probabilistic predictions since they minimize worst-case expected losses, and are included in the confidence intervals for any group and label. This enhanced reliability is corroborated by the experimental results of Section~\ref{sec:experimental_results}. Furthermore, the proposed methods can effectively leverage general \acp{LF} and exploit general information related to the \acp{LF}, as detailed in the following.

\section{Feature mapping components and expectation estimates for general \acp{LF}} \label{sec:uncertainty_sets}

Let $\Phi(\Lambda, y)\in\mathbb{R}^d$ be the feature mapping considered, its different components $\Phi^{(s)}$, for \mbox{$s=1,2,\ldots,d$}, allow us to encode multiple characteristics related to the \acp{LF}. Some components can describe the fit of the guesses provided by different \acp{LF}. In particular, if the $j$-th \ac{LF} guesses labels, components given by 
$$\Phi^{(s)}(L^{j}, y) = \mathbb{I}\{L^{j} \neq y\}$$
can encode the \ac{LF}'s error and components given by 
$$\Phi^{(s)}(L^{j}, y) = \mathbb{I}\{L^{j} \neq \mbox{`?'}\}$$ 
can encode the abstention probability. In addition, if the $j$-th \ac{LF} guesses label probabilities, components as 
$$\Phi^{(s)}(L^{j}, y) = (1 - L^{j}(y))^2$$ 
can encode the \ac{LF}'s Brier score and components as 
$$\Phi^{(s)}(L^{j}, y)=-\log (L^{j}(y))$$ 
can encode the \ac{LF}'s log score. 

Other components can describe the observed interdependencies of \acp{LF}. In particular, the disagreement between the $j$-th and $k$-th \acp{LF} can be encoded using 
    $$\Phi^{(s)}(L^{j}, L^{k}) = \mathbb{I}\{L^{j} \neq L^{k}\}$$
    for cases 
where both \acp{LF} guess labels, using 
    $$\Phi^{(s)}(L^{j}, L^{k}) = 1- L^{k}(L^{j})$$
    for cases where the $j$-th \ac{LF} guesses labels and the other guesses label probabilities,
or $$\Phi^{(s)}(L^{j}, L^{k}) = \|L^{j}- L^{k}\|$$ for cases 
where both guess label probabilities. Furthermore, the 
disagreement among more than two \acp{LF} can be encoded similarly, as well as that with respect to an expert predictor obtained from multiple \acp{LF}. 
Specifically, if $\text{MV}(\Lambda)$ denotes the \ac{MV} of the \acp{LF}' guesses $\Lambda=L^1,L^2,\ldots,L^m$, that is
$$\text{MV}(\Lambda) \in \arg\max_{y\in\set{Y}} \sum_{j = 1}^m\mathbb{I}\{L^j = y\}$$
its disagreement with  the $j$-th \ac{LF} can be encoded as $$\Phi^{(s)}(L^{j}, \Lambda) = 
\mathbb{I}\{L^{j} \neq \text{MV}(\Lambda)\}.$$

Expectation estimates $\widehat{\B{\tau}}\in\mathbb{R}^d$ for feature mappings can be directly obtained using the unlabeled dataset, prior knowledge, or a reduced set of labels. Specifically, the expectations corresponding to features that do not depend on labels (e.g., abstentions probabilities and disagreements) can be computed exactly just by evaluating the \acp{LF} in a large enough number of unlabeled instances. For example, the expectation $$\tau^{(s)}=\frac{1}{n}\sum_{i=1}^n\Phi^{(s)}(L^{j}_i, L^{k}_i)$$ can be computed by evaluating $L^{j}_i$ and $L^{k}_i$ for $i=1,2,\ldots n$, so that $\widehat{\tau}^{(s)}=\tau^{(s)}$ and $\lambda^{(s)}=0$. The estimates corresponding to features $\Phi^{(s)}(L^{j}, y)$ that encode \acp{LF}' errors or scores can be directly obtained from prior knowledge of the \acp{LF}' performance or by using a reduced set of labels. Specifically, the expectation $$\tau^{(s)}=\frac{1}{n}\sum_{i=1}^n\Phi^{(s)}(L^{j}_i, y_i)$$ can be estimated using labeled instances in a reduced subset~$\set{J}$ through the sample mean $$\widehat{\tau}^{(s)}=\frac{1}{|\set{J}|}\sum_{i\in\set{J}}\Phi^{(s)}(L^{j}_i, y_i)$$ and the corresponding $\lambda^{(s)}$ can be taken to be the standard error of such sample mean.

In practice, the feature components are chosen based on the \acp{LF}' types and the sources of information available. The next result shows that the usage of components given by disagreements also enables to encode characteristics for which we may not have explicit expectation estimates. %In particular, the next result shows how estimates for the error probability of the $j$-th \ac{LF} and the disagreement with the $k$-th \ac{LF} constraint the error probabilities for the $k$-th \ac{LF}.

\begin{theorem}\label{th:agreement}
The error probabilities and expected disagreement corresponding with the $j$-th and $k$-th \acp{LF} for \mbox{$j,k\in\{1,2,\ldots,m\}$} satisfy
\begin{align*}
& \left|\frac{1}{n}\sum_{i = 1}^n \mathbb{I}\{L_{i}^{j}\neq y_i\}  - \frac{1}{n}\sum_{i = 1}^n \mathbb{I}\{L_{i}^{k}\neq y_i\}\right|\leq \sum_{i = 1}^n \frac{\mathbb{I}\{L_{i}^{j}\neq L_{i}^{k}\}}{n}.
\end{align*}
% \vspace{-0.4cm}
\begin{proof}
   The result is obtained since the disagreement between pairs of \acp{LF} satisfies the triangle inequality, that is for any two \acp{LF} $L^j$ and $L^k$ we have that
  \begin{align}
      \label{eq:triangle_inequality}
&\frac{1}{n}\sum_{i = 1}^n \mathbb{I}\{L_{i}^{j}\neq y_i\} \leq \frac{1}{n}\sum_{i = 1}^n \mathbb{I}\{L_{i}^{k}\neq y_i\} + \frac{1}{n}\sum_{i = 1}^n \mathbb{I}\{L_{i}^{j}\neq L_{i}^{k}\}.
\end{align} 
where $\frac{1}{n}\sum_{i = 1}^n \mathbb{I}\{L_{i}^{j}\neq L_{i}^{k}\}$ is the disagreement between $L^j$ and $L^k$, and $\frac{1}{n}\sum_{i = 1}^n \mathbb{I}\{L_{i}^{j}\neq y_i\}$ and $\frac{1}{n}\sum_{i = 1}^n \mathbb{I}\{L_{i}^{k}\neq y_i\}$ are the errors of $L^j$ and $L^k$, respectively. Such inequality in \eqref{eq:triangle_inequality} is directly obtained because 
 $$\{L^j \neq L^k\} \subseteq \{L^j \neq y\} \cup \{L^k\neq y\}$$ 
since if the $j$-th and $k$-th LFs provide different outputs, at least one of them have to provide an output different to $y$. 

Therefore, we have that 
  \begin{align*}
& \left|\frac{1}{n}\sum_{i = 1}^n \mathbb{I}\{L_{i}^{j}\neq y_i\} - \frac{1}{n}\sum_{i = 1}^n \mathbb{I}\{L_{i}^{k}\neq y_i\} \right|\leq \sum_{i = 1}^n \frac{\mathbb{I}\{L_{i}^{j}\neq L_{i}^{k}\}}{n}
\end{align*}
because inequality in \eqref{eq:triangle_inequality} implies 
  \begin{align*}
& \frac{1}{n}\sum_{i = 1}^n \mathbb{I}\{L_{i}^{j}\neq y_i\} - \frac{1}{n}\sum_{i = 1}^n \mathbb{I}\{L_{i}^{k}\neq y_i\} \leq \frac{1}{n}\sum_{i = 1}^n \mathbb{I}\{L_{i}^{j}\neq L_{i}^{k}\}
\end{align*} 
and, reversing the roles of $j$-th and $k$-th \acp{LF} in \eqref{eq:triangle_inequality}, we have that
  \begin{align*}
\frac{1}{n}\sum_{i = 1}^n \mathbb{I}\{L_{i}^{k}\neq y_i\} - \frac{1}{n}\sum_{i = 1}^n \mathbb{I}\{L_{i}^{j}\neq y_i\}\leq \frac{1}{n}\sum_{i = 1}^n \mathbb{I}\{L_{i}^{j}\neq L_{i}^{k}\}.
\end{align*}
\end{proof}
% \vspace{-0.3cm}
%% \vspace{-00.3cm}
\end{theorem}

The theorem above shows that estimates for expected disagreements together with estimates for error probabilities also serve to constrain other error probabilities.
Such theorem is stated in terms of pairs of \acp{LF} but analogous results are satisfied in other cases. For instance, estimates for the error probability of the \ac{MV} and its disagreement with individual \acp{LF} also constrain the errors of the \acp{LF} even if they are not explicitly estimated. 

The proposed methodology can utilize assorted types of feature mappings depending on the prior knowledge avaible and the outputs of the LFs. For instance, if a set of labeled instances is available and the outputs of the LFs are labels, it is appropriate to use feature mappings that encode the error behavior of the LFs (e.g., $\Phi^{(s)}(L^{j}, y) = \mathbb{I}\{L^{j} \neq y\}$), as the expectation can be estimated from the labeled data. In contrast, if no labeled data is available and the outputs of the LFs are probabilities, it is appropriate to use feature mappings based on disagreements between pairs of LFs (e.g., $\Phi^{(s)}(L^{j}, L^{k}) = \|L^{j}- L^{k}\|$), as the expectation can be computed just from the outputs of the LFs. 

\section{Experimental results}\label{sec:experimental_results}

This section assesses the performance of the methods proposed using multiple datasets. In the first set of numerical results, we show the reliability of the presented confidence intervals for label probabilities; the second set of numerical results compares the performance of the proposed \ac{WMRC} with respect to that of existing techniques; and the third set of numerical results further evaluate the probabilistic performance of the proposed \ac{WMRC}. The code used in the
experimental results is provided on the web \url{https://github.com/MachineLearningBCAM/Weak-Supervision-TPAMI-2025}.   In addition, the running time of the proposed method is in the order of 30 seconds per dataset. % In the supplementary materials, we provide the code for the methods proposed and Appendix~\ref{app:numerical_results} shows additional numerical results that further describe the reliability of the presented confidence intervals and probabilistic predictions.

The proposed method is evaluated using 8 publicly available datasets: \ac{awa} \cite{xian2018zero, Mazzetto2021}, DomainNet (Domain) \cite{Mazzetto2021}, IMDB \cite{ren2021denoising, zhang2021wrench}, Yelp \cite{ren2021denoising, zhang2021wrench}, Basketball (Basket) \cite{caba2015activitynet, Fu2020}, SMS \cite{awasthi2020Learning, zhang2021wrench}, OBS Network~\cite{Arachie2019}, and Cardiotocography (Cardio) \cite{Arachie2019}. The \acp{LF} of the first 2 datasets provide labels, the \acp{LF} of the last 2 datasets provide probabilities, and the \acp{LF} of the rest of the datasets provide labels and abstentions.

\begin{figure}
\centering
\psfrag{Probabilities}[][][0.8]{Probability}
\psfrag{x}[r][r][0.7]{r}
\psfrag{data40}[l][l][0.8]{MMP $y = 0$}% Pr. prediction $y = 0$}
\psfrag{data15}[l][l][0.8]{MMP $y = 1$}% Pr. prediction $y = 1$}
\psfrag{data13}[l][l][0.8]{MMP $y = 0$}% Pr. prediction $y = 0$}
\psfrag{data33}[l][l][0.8]{MMP $y = 1$}% Pr. prediction $y = 1$}
\psfrag{data12}[l][l][0.8]{Actual probability}
\psfrag{data10}[l][l][0.8]{MMP $y = 1$}% Pr. prediction $y = 1$}
\psfrag{data20}[l][l][0.8]{MMP $y = 0$}% Pr. prediction $y = 0$}
\psfrag{data39}[l][l][0.8]{MMP $y = 1$}% Pr. prediction $y = 1$}
\psfrag{data1}[l][l][0.8]{MMP $y = 0$}% Pr. prediction $y = 0$}
\psfrag{data43}[l][l][0.8]{Actual probability}
\psfrag{data19}[l][l][0.8]{Actual probability}
\psfrag{Confidence intervalabcdefghijklmno}[l][l][0.8]{Confidence interval}
\psfrag{Confidence intervalabcdefghijklmnopqrstuv}[l][l][0.8]{Confidence interval}
\psfrag{data46}[l][l][0.8]{Confidence interval $\delta = 0.05$}
\psfrag{data17}[l][l][0.8]{Actual probability}
\psfrag{data3}[l][l][0.8]{Actual probability}
\psfrag{r}[l][l][0.8]{$p$}
\psfrag{p}[l][l][0.8]{$r$}
\psfrag{3}[r][r][0.7]{28/36}
\psfrag{7}[r][r][0.7]{21/36}
\psfrag{11}[r][r][0.7]{13/36}
\psfrag{19}[r][r][0.7]{12/36}
\psfrag{23}[r][r][0.7]{9/38}
\psfrag{26}[r][r][0.7]{28/36}
\psfrag{30}[r][r][0.7]{24/36}
\psfrag{34}[r][r][0.7]{20/36}
\psfrag{42}[r][r][0.7]{17/36}
\psfrag{46}[r][r][0.7]{9/36}
\psfrag{1}[r][r][0.7]{0.95}
\psfrag{5}[r][r][0.7]{.75}
\psfrag{10}[r][r][0.7]{.90}
\psfrag{14}[r][r][0.7]{.70}
\psfrag{18}[r][r][0.7]{.50}
\psfrag{15}[r][r][0.7]{.75}
\psfrag{20}[r][r][0.7]{.90}
\psfrag{0.4}[r][r][0.7]{.4}
\psfrag{0.6}[r][r][0.7]{.6}
\psfrag{0.8}[r][r][0.7]{.8}
\psfrag{0.7}[r][r][0.7]{.7}
\psfrag{0.85}[r][r][0.7]{.85}
\psfrag{0.9}[r][r][0.7]{.9}
\psfrag{0.95}[r][r][0.7]{.95}
\psfrag{2}[r][r][0.7]{.85}
\psfrag{6}[r][r][0.7]{.65}
\psfrag{1}[r][r][0.7]{1}
% \vspace{-0.3cm}
\includegraphics[width=0.48\textwidth]{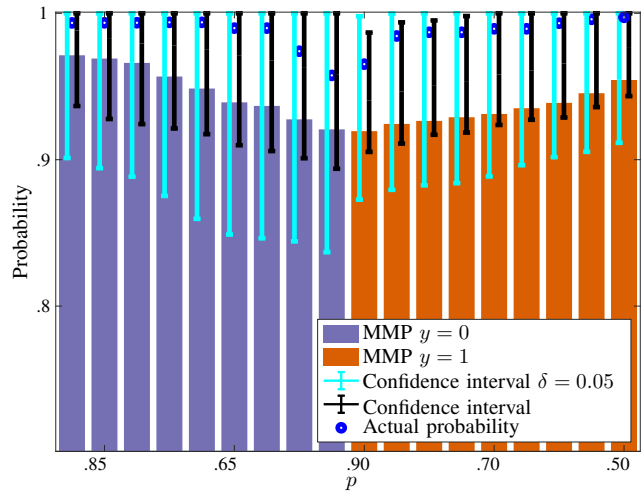}
\caption{% Confidence intervals and MMP probabilities grouping 
Results on \ac{awa} dataset show the reliability of the presented confidence intervals and probabilistic predictions grouping instances as in~\eqref{eq:groups}.}
\label{fig:groups}
\end{figure}

The proposed methodology is compared with the baseline \ac{MV} and 5 techniques for programmatic weak supervision: \ac{AMCL} \cite{Mazzetto2021}, FlyingSquid~\cite{Fu2020}, \ac{EBCC} \cite{li2019exploiting}, hyper label model \cite{wulearning}, and Snorkel~\cite{Ratner2017}. All the numerical results are obtained averaging the predictions achieved in~10 random instantiations, and the hyper-parameters in FlyingSquid, \ac{EBCC}, hyper label model, and Snorkel are set to the default values provided by the Wrench library \cite{zhang2021wrench} and provided by the authors.\footnote{For the \ac{AMCL} method the scaling factor used is 0.4 in order to avoid infeasible optimization problems that occurred with the value 0.1 suggested by the authors, while both scaling factors provide similar results in cases with feasible optimizations.} \ac{WMRC} probabilities are obtained by using feature mappings that can encode the \acp{LF}' error ($\Phi^{(s)}(L^{j}, y) = \mathbb{I}\{L^{j} \neq y\}$), abstention probabilities ($\Phi^{(s)}(L^{j}, y) = \mathbb{I}\{L^{j} \neq \mbox{`?'}\}$), the probabilities' Brier score ($\Phi^{(s)}(L^{j}, y)= (1-L^{j}(y))^2$), disagreements between \acp{LF} and the \ac{MV} \mbox{($\Phi^{(s)}(L^{j}, \Lambda) = 
\mathbb{I}\{L^{j} \neq \text{MV}(\Lambda)\}$)} and disagreements between pairs of \acp{LF} that guess label probabilities ($\Phi^{(s)}(L^{j}, L^{k}) = \|L^{j}- L^{k}\|$). Expectation estimates for components that do not depend on labels are computed exactly as described in Section~\ref{sec:uncertainty_sets}, while expectation estimates for feature mappings that depend on labels are obtained using estimates of the errors of the \acp{LF}. Specifically, the methods \ac{WMRC} and \ac{AMCL} \cite{Mazzetto2021,mazzetto2021semi} estimate the errors of the \acp{LF} using $100$ labeled instances randomly sampled.   In practice, the uncertainty set may become empty.  These cases can be easily detected (unbounded objective), and can be avoided by increasing some components of $\B{\lambda}$ to ensure the uncertainty set contains some distributions. For instance, such vector can be given by $\B{\tilde{\lambda}} = \max(\B{\lambda}, |\B{\widehat{\tau}} - \B{\tilde{\tau}}|)$ with$${\tilde{\tau}}^{(s)}=\frac{1}{n} \sum_{i = 1}^{n}\sum_{y\in\set{Y}} \mathbb{I}\{\text{MV}(\Lambda_i)=y\} \Phi(\Lambda_i,y)$$
for $s = 1, 2, \ldots, d$.  The usage of such increased vector  $\tilde{\B{\lambda}}$ ensures that probabilities  $\up{p}_{i}(y) = \mathbb{I}\{\text{MV}(\Lambda_i)=y\}$ for \mbox{$i = 1, 2, \ldots, n$} are in the uncertainty set. % The vector $\tilde{\B{\tau}}$ is given by the agreement between each LF and the MV. % This ensures that the uncertainty set determined by $\tilde{\B{\tau}}$ and $\tilde{\B{\lambda}}$ contains the probabilities $\mathbb{I}\{\text{MV}(\Lambda_i)=y\}$. Then, this uncertainty set is not empty.

\begin{figure*}
\centering
\psfrag{Probabilities}[][][0.8]{Probability}
\psfrag{x}[r][r][0.8]{r}
\psfrag{data25}[l][l][0.8]{Actual probability}
\psfrag{data24}[l][l][0.8]{Confidence interval}
\psfrag{r}[l][l][0.7]{$p$}
\psfrag{p}[l][l][0.8]{$r$}
\psfrag{27}[][][0.8]{27/36}
\psfrag{25}[][][0.8]{}%25/36}
\psfrag{23}[][][0.8]{23/36}
\psfrag{21}[][][0.8]{}%21/36}
\psfrag{19}[][][0.8]{19/36}
\psfrag{17}[][][0.8]{}%17/36}
\psfrag{15}[][][0.8]{15/36}
\psfrag{13}[][][0.8]{}%13/36}
\psfrag{11}[][][0.8]{11/36}
\psfrag{9}[][][0.8]{}%9/36}
\psfrag{1}[r][r][0.7]{0.95}
\psfrag{5}[r][r][0.7]{0.75}
\psfrag{10}[r][r][0.7]{0.90}
\psfrag{14}[r][r][0.7]{0.70}
\psfrag{18}[r][r][0.7]{0.50}
\psfrag{20}[r][r][0.7]{0.90}
\psfrag{0.4}[r][r][0.7]{.4}
\psfrag{0.6}[r][r][0.7]{.6}
\psfrag{0.8}[r][r][0.7]{.8}
\psfrag{0.7}[r][r][0.7]{.7}
\psfrag{0.85}[r][r][0.7]{0.85}
\psfrag{0.9}[r][r][0.7]{0.9}
\psfrag{0.95}[r][r][0.7]{0.95}
\psfrag{2}[r][r][0.7]{0.85}
\psfrag{6}[r][r][0.7]{0.65}
\psfrag{1}[r][r][0.7]{1}
\psfrag{r}[l][l][0.8]{$r$}
% \subfigure[Confidence intervals and \mbox{MP} probabilities grouping instances as in~\eqref{eq:groups_LFs}\label{fig:groups_LFs}]{ \includegraphics[width=0.32\textwidth]{LFs_agree_2classes_2.eps}}\hfill
\psfrag{4}[r][r][0.7]{21/36}
\psfrag{12}[r][r][0.7]{25/36}
\psfrag{8}[r][r][0.7]{13/36}
\psfrag{16}[r][r][0.7]{12/36}
\psfrag{20}[r][r][0.7]{9/36}
\psfrag{data1}[l][l][0.8]{AMCL $y = 0$}
\psfrag{data20}[l][l][0.8]{AMCL $y = 1$}
\psfrag{Confidence intervalabc}[l][l][0.8]{MMP $y = 0$}
\psfrag{data16}[l][l][0.8]{MMP $y = 1$}
\psfrag{data22}[l][l][0.8]{Confidence interval}
\psfrag{data23}[l][l][0.8]{Actual probability}
% \vspace{0.1cm}
\psfrag{data4}[l][l][0.8]{Snorkel $y = 0$}
         \subfloat[\small Cases with class $y = 0$]{\includegraphics[width=0.48\textwidth]{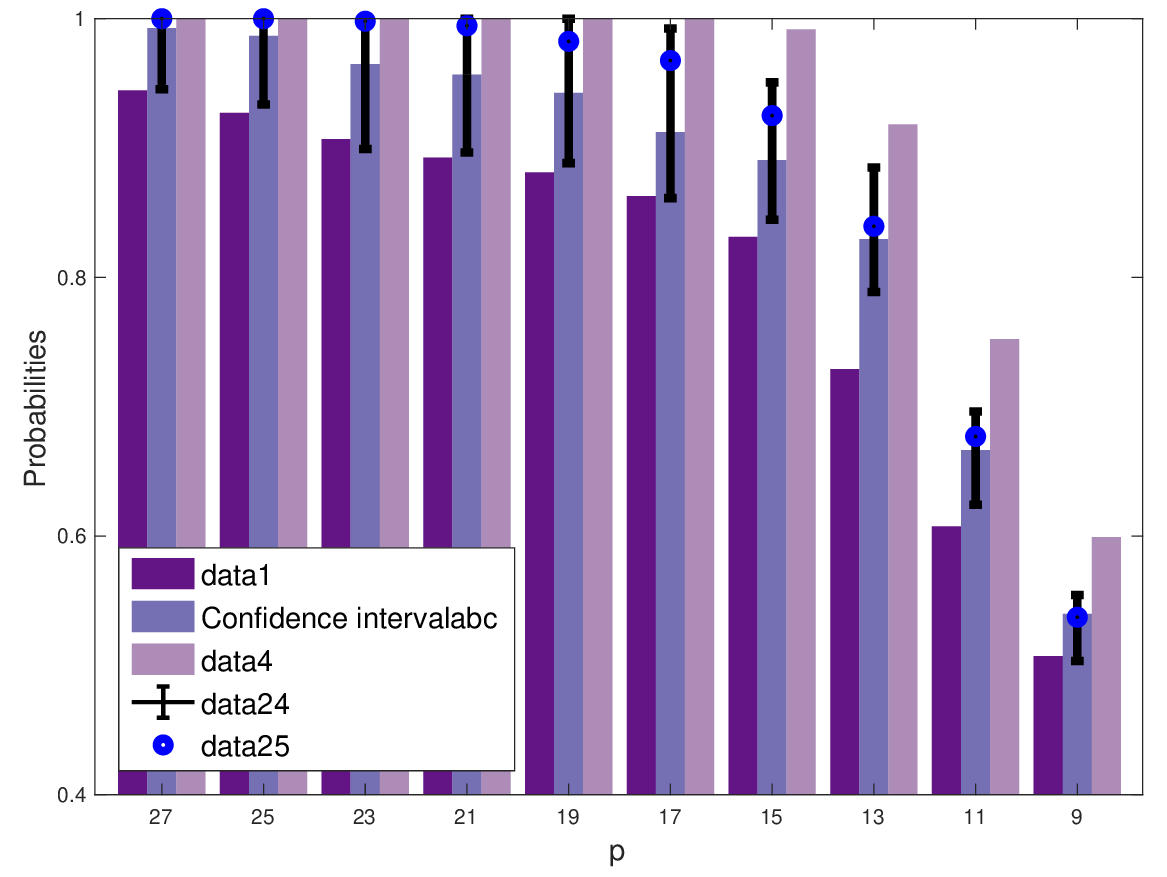}}
        \psfrag{data4}[l][l][0.8]{Snorkel $y = 1$}
        \psfrag{Confidence intervalabcdefghiklm}[l][l][0.8]{MMP $y = 1$}
         \subfloat[\small Cases with class $y = 1$]{\includegraphics[width=0.48\textwidth]{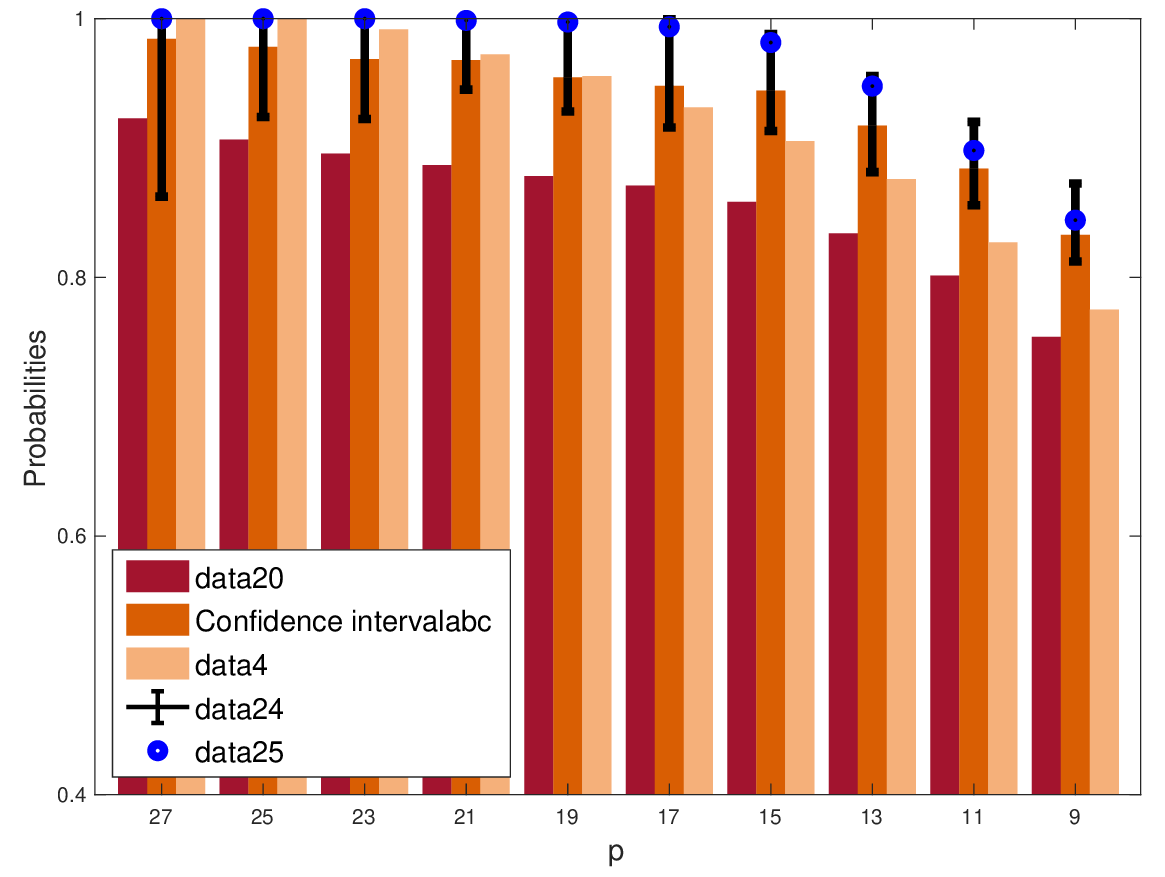}}
         \caption{% Confidences intervals, MMP, and AMCL\_CC probabilities grouping 
         Results on \ac{awa} dataset show the reliability of the presented confidence intervals and probabilistic predictions grouping instances as in~\eqref{eq:groups_LFs}.}
                  \label{fig:groups_LFs2}
\end{figure*}

In the first set of numerical results, we show the reliability of the presented confidence intervals and the probabilistic predictions using \acf{awa} dataset. 
For different groups and labels, Figures~\ref{fig:groups} and~\ref{fig:groups_LFs2} show the proposed confidence intervals given by the optimal values of the optimization problems in~\eqref{eq:confidence_interval_upper}-\eqref{eq:confidence_interval_lower} in comparison with the actual probabilities $\up{p}_\set{I}^*(y)$ and probabilistic predictions $\up{h}_\set{I}(y)$ given by \ac{WMRC} and \ac{AMCL} methods. 
In particular, Figure~\ref{fig:groups} shows confidence intervals grouping instances based on the predicted probabilities as in~\eqref{eq:groups} with $p$ given by $\{0.9, 0.85, 0.8, 0.75, 0.7, 0.65, 0.6, 0.55, 0.5\}$. In that figure, we use $\B{\lambda}$ given by the Wilson’s interval for a binomial proportion with coverage probability $1-\delta = 0.95$ (light blue error bar) and $\B{\lambda}$ given by standard sample errors (black error bar). This figure shows that the proposed confidence intervals can provide informative and practically useful bounds for actual label probabilities, even using approximate assessments for estimation errors. In particular, the confidence intervals obtained using standard errors contain the actual and \ac{WMRC} probabilities in all the experimental results carried out.% (see also Appendix~\ref{app:numerical_results}).

Figure~\ref{fig:groups_LFs2} shows the confidence intervals grouping instances based on the outputs of \acp{LF} as in~\eqref{eq:groups_LFs} with $r$ given by $\{27/36, 25/36, 23/36, \ldots, 9/36\}$. % in comparison with the actual and \ac{WMRC} probabilities. 
Such figure further shows that the confidence intervals reliably assess the actual probabilities and that \ac{WMRC} probabilities are often close to the actual probabilities. In particular, both actual and \ac{WMRC} probabilities are contained in the confidence intervals in agreement with the theoretical results in Theorems~\ref{th:confidence_interval} and~\ref{th:prob_estimated}. Figure~\ref{fig:groups_LFs2} also shows that the methods presented can provide more reliable probabilistic predictions than the state-of-the-art. The figure compares \ac{WMRC} predictions with those of \ac{AMCL} and Snorkel methods. % for a fair comparison since both methods are implemented using few labeled samples. 
Table~\ref{tab:error_transductive} below further quantifies the probabilistic performance improvement of the proposed method in comparison with the other existing techniques.

\begin{table*}
\begin{center}
\caption{Average Brier's score, average calibration error, average log-loss, and average 0-1 loss  $\pm$ standard deviation of the proposed method in comparison with existing techniques.\vspace{0.1cm}}
\label{tab:error_transductive}
\renewcommand{\arraystretch}{1.2}
% %% \vspace{-00.2cm}
\begin{adjustbox}{width=\textwidth,center}
% %% \vspace{-00.5cm}
\begin{tabular}{c|l|rrrrrrrrrrrrrrrrrrrrr}
\toprule 
Score & Method                                 & \multicolumn{1}{c}{Awa \cite{xian2018zero}} & \multicolumn{1}{c}{OBS \cite{Arachie2019}} & \multicolumn{1}{c}{Cardio \cite{Arachie2019}} &  \multicolumn{1}{c}{IMDB \cite{ren2021denoising}}  & \multicolumn{1}{c}{Yelp \cite{ren2021denoising}} & \multicolumn{1}{c}{Basket \cite{Fu2020, caba2015activitynet}} &  \multicolumn{1}{c}{SMS \cite{awasthi2020Learning}} &  \multicolumn{1}{c}{Domain \cite{Mazzetto2021}} \\ 
 \hline 
\multirow{7}{*}{\rotatebox[origin=c]{90}{Brier score}} & \ac{MV}  & .07$\pm$.00& .30$\pm$.00& .19$\pm$.00& .24$\pm$.00& .23$\pm$.00& .20$\pm$.00&  .22$\pm$.00 & .62$\pm$.00 \\
& Snorkel$^{\footnotesize 2}$ \cite{Ratner2017}  &.06$\pm$.01 & .29$\pm$.00 & .26$\pm$.00 & \textbf{.20}$\pm$.00 & \textbf{.19}$\pm$.00 & \textbf{.10}$\pm$.00  & .13$\pm$.00 & .68$\pm$.00 \\
& FlyingSquid$^{\footnotesize 2}$ \cite{Fu2020} & \textbf{.02}$\pm$.00& .27$\pm$.00& .26$\pm$.00& .31$\pm$.00& .43$\pm$.00& .22$\pm$.00& .14$\pm$.00& .83$\pm$.00\\
& EBCC$^{\footnotesize 2}$ \cite{li2019exploiting} & .04$\pm$.00& .31$\pm$.00&  .30$\pm$.00& .23$\pm$.00& .24$\pm$.00&.18$\pm$.00& .16$\pm$.00& .74$\pm$.00\\
& HyperLM$^{\footnotesize 2}$ \cite{wulearning} & .05$\pm$.00& {.25}$\pm$.00&  .16$\pm$.00& .21$\pm$.00& .20$\pm$.00& .27$\pm$.00& .30$\pm$.00& .65$\pm$.00\\
& AMCL$^{\footnotesize 2, 3}$ \cite{Mazzetto2021,mazzetto2021semi} & .03$\pm$.01 & .27$\pm$.01 & \textbf{.03}$\pm$.01& .23$\pm$.02 & .24$\pm$.02 & .11$\pm$.01 & .25$\pm$.03 & \textbf{.54}$\pm$.01 \\
& MMP & \textbf{.02}$\pm$.01  & \textbf{.23}$\pm$.02 & .04$\pm$.01 & \textbf{.20}$\pm$.01 & .20$\pm$.06 & .11$\pm$.01 & \textbf{.12}$\pm$.01 & .55$\pm$.02 \\
\hline
\multirow{7}{*}{\rotatebox[origin=c]{90}{Calibration error}}& MV                                                          & .26$\pm$.00                                                   & .30$\pm$.00                                                  & .26$\pm$.00                                                     & .33$\pm$.00                                                        & .36$\pm$.00                                                        & .24$\pm$.00                                                & .32$\pm$.00                                                          & .24$\pm$.00                                                      \\
& Snorkel    \cite{Ratner2017}                                                 & .23$\pm$.02                                                   & .27$\pm$.01                                                  & .39$\pm$.01                                                     & .19$\pm$.02                                                        & .40$\pm$.04                                                        & \textbf{.14$\pm$.01}                                       & .34$\pm$.03                                                          & .14$\pm$.00                                                      \\
& FlyingSquid    \cite{Fu2020}                                             & \textbf{.07$\pm$.00}                                          & .27$\pm$.00                                                  & .39$\pm$.00                                                     & .46$\pm$.00                                                        & .45$\pm$.00                                                        & .30$\pm$.00                                                & .40$\pm$.00                                                          & .22$\pm$.00                                                      \\
& EBCC        \cite{li2019exploiting}                                                & .27$\pm$.00                                                   & .27$\pm$.00                                                  & .42$\pm$.00                                                     & \textbf{.18$\pm$.00}                                               & .42$\pm$.00                                                        & .26$\pm$.00                                                & .38$\pm$.00                                                          & .24$\pm$.00                                                      \\
& HyperLM    \cite{wulearning}                                                 & .18$\pm$.00                                                   & .41$\pm$.00                                                  & .39$\pm$.00                                                     & .24$\pm$.00                                                        & .47$\pm$.00                                                        & .36$\pm$.00                                                & .42$\pm$.00                                                          & .21$\pm$.00                                                      \\
& AMCL      \cite{Mazzetto2021,mazzetto2021semi}                                                  & .13$\pm$.01                                                   & .29$\pm$.01                                                  & \textbf{.08$\pm$.02}                                            & .47$\pm$.03                                                        & .38$\pm$.03                                                        & .20$\pm$.03                                                & .39$\pm$.01                                                          & \textbf{.10$\pm$.01}                                             \\
& MMP                                                         & \textbf{.07$\pm$.02}                                          & \textbf{.20$\pm$.04}                                         & \textbf{.08$\pm$.02}                                            & .21$\pm$.04                                                        & \textbf{.24$\pm$.02}                                               & .20$\pm$.06                                                & \textbf{.30$\pm$.02}                                                 & \textbf{.10$\pm$.01}                                             \\ \hline
\multirow{7}{*}{\rotatebox[origin=c]{90}{Log-loss}}& {MV} & .31$\pm$.00  & 8.73$\pm$.00 & .66$\pm$.00 & 6.39$\pm$.00 & 5.90$\pm$.00 & 2.40$\pm$.00 & .79$\pm$.00 & 5.48$\pm$.00 \\
& Snorkel \cite{Ratner2017} & .42$\pm$.02&3.98$\pm$.01&7.01$\pm$.01&{.68}$\pm$.02&2.61$\pm$.04&1.31$\pm$.01& {.53}$\pm$.03& 9.21$\pm$.00\\
& FlyingSquid \cite{Fu2020}&\textbf{.11$\pm$.00} & 2.23$\pm$.00 & .85$\pm$.00 & .82$\pm$.00 & 1.34$\pm$.00 & \textbf{.39$\pm$.00} & .73$\pm$.00 & 2.01$\pm$.00 \\
& EBCC \cite{li2019exploiting}&.13$\pm$.00 &2.23$\pm$.00 &.91$\pm$.00&.73$\pm$.00&.81$\pm$.00& .45$\pm$.00  & .43$\pm$.00 & 1.80$\pm$.00 \\
& HyperLM \cite{wulearning} & .21$\pm$.00 &2.66$\pm$.00&.60$\pm$.00&.62$\pm$.00&\textbf{.60$\pm$.00}& 1.31$\pm$.00 & .68$\pm$.00 & 1.29$\pm$.00 \\
& AMCL \cite{Mazzetto2021,mazzetto2021semi} & .15$\pm$.01 &{8.73}$\pm$.01&.40$\pm$.01&1.47$\pm$.02&.82$\pm$.03&{1.52}$\pm$.03 &.69$\pm$.01& 5.43$\pm$.01\\
& MMP &\textbf{.11$\pm$.02}&\textbf{.61$\pm$.03} &\textbf{.21$\pm$.04} & \textbf{.60$\pm$.02} & \textbf{.60$\pm$.01}  & \textbf{.39$\pm$.05}   & \textbf{.41$\pm$.02}  & \textbf{1.11$\pm$.01}     \\
\hline
  % & \multicolumn{1}{c}{CE} & \multicolumn{1}{c}{BS} & \multicolumn{1}{c}{CE} & \multicolumn{1}{c}{BS} & \multicolumn{1}{c}{CE} & \multicolumn{1}{c}{BS}& \multicolumn{1}{c}{CE} & \multicolumn{1}{c}{BS}& \multicolumn{1}{c}{CE} & \multicolumn{1}{c}{BS}& \multicolumn{1}{c}{CE} & \multicolumn{1}{c}{BS}& \multicolumn{1}{c}{CE} & \multicolumn{1}{c}{BS}    & \multicolumn{1}{c}{CE} & \multicolumn{1}{c}{BS} \\ \hline
\multirow{7}{*}{\rotatebox[origin=c]{90}{0-1 loss}} & \ac{MV} & \textbf{.01}$\pm$.00&.28$\pm$.00& .35$\pm$.00& .29$\pm$.00& .32$\pm$.00& .25$\pm$.00& .32$\pm$.00&.46$\pm$.00\\
& Snorkel \cite{Ratner2017} & .03$\pm$.01 &.28$\pm$.00  & .40$\pm$.00 &.30$\pm$.00&.47$\pm$.00&\textbf{.11}$\pm$.00 &.32$\pm$.00  &.73$\pm$.00 \\
& FlyingSquid \cite{Fu2020} & \textbf{.01}$\pm$.00& .38$\pm$.00& .40$\pm$.00& .30$\pm$.00& .33$\pm$.00& .35$\pm$.00& .13$\pm$.00& .40$\pm$.00\\
& EBCC \cite{li2019exploiting} & .03$\pm$.00& .28$\pm$.00& .40$\pm$.00& \textbf{.28}$\pm$.00& .36$\pm$.00& .36$\pm$.00&\textbf{.08}$\pm$.00& .48$\pm$.00\\
& HyperLM \cite{wulearning} & .03$\pm$.00& .28$\pm$.00& .08$\pm$.00&\textbf{.28}$\pm$.00& .33$\pm$.00& .36$\pm$.00& .54$\pm$.00& .42$\pm$.00\\
& AMCL \cite{Mazzetto2021,mazzetto2021semi} & .02$\pm$.01  & .28$\pm$.01 &\textbf{.03}$\pm$.01 & .31$\pm$.02  & .38$\pm$.02 & .12$\pm$.01 & .45$\pm$.03 & \textbf{.37}$\pm$.01 \\
& MMP & \textbf{.01}$\pm$.01 & \textbf{.27}$\pm$.02& \textbf{.03}$\pm$.01 & .29$\pm$.01 & \textbf{.28}$\pm$.06 & \textbf{.11}$\pm$.01&\textbf{.08}$\pm$.01 &\textbf{.37}$\pm$.02 \\
\bottomrule
\end{tabular}
\end{adjustbox}
% \vspace{-0.3cm}
\end{center}
$^{\footnotesize 2}$\footnotesize{ Methods that do not allow probabilistic predictions of the LFs. $^{\footnotesize 3}$ Methods that do not allow abstentions.}
\end{table*}

\begin{figure}
\centering
\psfrag{Probability}[b][][0.8]{Probability}
\psfrag{Number of LFs}[t][][0.8]{Number of LFs}
\psfrag{Confidence intervalabc}[l][l][0.8]{MMP $y = 0$}
\psfrag{data2}[l][l][0.8]{MMP $y = 1$}
\psfrag{data3}[l][l][0.8]{Confidence interval}
\psfrag{data4}[l][l][0.8]{Actual probability}
\psfrag{11}[][][0.7]{11}
\psfrag{19}[][][0.7]{19}
\psfrag{27}[][][0.7]{27}
\psfrag{35}[][][0.7]{35}
\psfrag{0.6}[][][0.7]{.6}
\psfrag{0.8}[][][0.7]{.8}
\psfrag{1}[][][0.7]{1}
% \vspace{-0.3cm}
\includegraphics[width=0.48\textwidth]{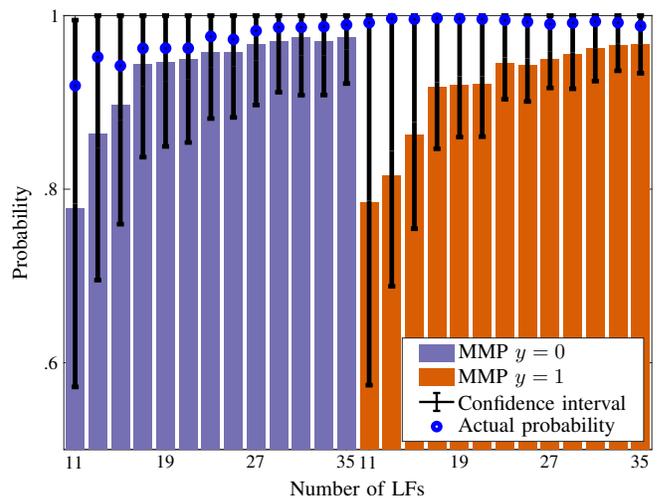}
\caption{% Confidence intervals and MMP probabilities grouping 
Results on 'AWA' dataset show the reliability of the presented confidence intervals and probabilistic predictions grouping instances as in~\eqref{eq:groups} varying the number of LFs.}
\label{fig:groups_increasingLF}
\end{figure}

Figure~\ref{fig:groups_increasingLF} shows the confidence intervals grouping instances based on the predicted probabilities as in~\eqref{eq:groups} with $p$ given by~$0.7$ and using $\B{\lambda}$ given by the standard sample errors. The figure shows how confidence intervals change for different number of LFs, in particular the addition of LFs results in tighter confidence intervals and more reliable probabilistic predictions. Furthermore, the figure shows that the proposed methods can provide  informative and practically useful bounds for actual label probabilities, even using a small number of~LFs.

In the second set of numerical results, we compare the performance of \ac{WMRC} with that of existing techniques. 
Table~\ref{tab:error_transductive} shows the average Brier score, calibration error, log-loss, and 0-1 loss using multiple datasets. The calibration error \cite{kumar2019verified} is a widely used metric to evaluate probabilistic predictions that assess the difference between predicted label probabilities and actual label proportions. Calibration error is quantified in Table~\ref{tab:error_transductive} using the efficient the method proposed in~\cite{kumar2019verified}. 
The results of the methods that do not allow abstentions of the \acp{LF} are obtained by replacing the abstentions by random labels, while the results of the methods that do not allow probabilistic predictions of the \acp{LF} are obtained by replacing the probabilities by the label with the highest probability. 

% Table~\ref{tab:error_transductive} shows that existing techniques such as EBCC \cite{li2019exploiting} can achieve a low prediction error in certain datasets (e.g., `SMS'), however their probabilistic predictions are often less reliable.  The table shows that the proposed techniques offer an overall improved performance in terms of Brier's score, calibration error, log-loss, and 0-1 loss (prediction error). Specifically, the \ac{WMRC} predictions provided by the presented methods rank among the top two results in all the datasets, both in terms of probabilistic scores and prediction errors. 

Table~\ref{tab:error_transductive} shows that the proposed methodology outperforms other methods in terms of four different and common metrics to assess probabilistic predictions. Some techniques can obtain slightly better performance than the proposed method in few datasets and in terms of some metric but the presented method provides an overall clear performance improvement with respect to all the other methods. For instance,  Snorkel method is slightly better than the presented MMP in terms of 0-1 loss in `Yelp' dataset, but is much worse than MMP in that dataset in all the other three metrics. The performance improvement of the presented methods in the four metrics shows the superiority of the proposed probabilistic predictions.

\begin{figure}
\centering
         \psfrag{Abstention threshold}[][b][0.8]{Abstention threshold}
         \psfrag{Snorkel}[l][l][0.8]{Snorkel}
         \psfrag{hyper labelabc}[l][l][0.8]{Hyper LM}
         \psfrag{EBCC}[l][l][0.8]{EBCC}
         \psfrag{WMRC}[l][l][0.8]{MMP}
         \psfrag{WMRC Majority voteabcde}[l][l][0.8]{WMRC-MV}
         \psfrag{WMRC exp}[l][l][0.8]{WMRC-dis}
         \psfrag{ACML}[l][l][0.8]{AMCL}
          \psfrag{0.5}[r][r][0.7]{.5}
         \psfrag{0.6}[r][r][0.7]{.6}
         \psfrag{0.7}[r][r][0.7]{.7}
          \psfrag{0.8}[r][r][0.7]{.8}
          \psfrag{0.9}[r][r][0.7]{.9}
          \psfrag{0}[r][r][0.7]{0}
                     \psfrag{Brier score}[][t][0.8]{Brier score}
          \psfrag{0.02}[r][r][0.7]{.02}
          \psfrag{0.04}[r][r][0.7]{.04}
          \psfrag{0.06}[r][r][0.7]{.06}
          \psfrag{0.08}[r][r][0.7]{.08}
\includegraphics[width=0.48\textwidth]{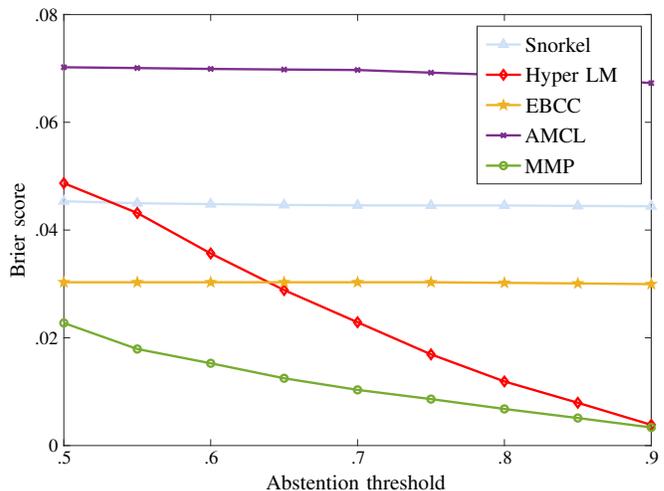}
 \caption{Results on \ac{awa} dataset show the reliability of the presented probabilistic predictions}
 \label{fig:abstain_brier}
\end{figure}

In the third set of numerical results, we further evaluate the probabilistic performance of the proposed methods and existing techniques.  Figure~\ref{fig:abstain_brier} shows the Brier score increasing the threshold used to decide when to abstain using '\ac{awa}' dataset. Specifically, the results are obtained computing the Brier score of instances where probabilities are above the threshold. Figure~\ref{fig:abstain_brier} shows that the Brier score of the \ac{WMRC} methods decreases by increasing the abstention threshold, so that abstaining in instances where probabilities are low ensures that decisions are only made when the model is confident.

The experimental results show that the proposed confidence intervals can offer informative upper and lower bounds for the actual probabilities, and contain the probabilities predicted by the presented methods. In addition, the numerical results show that the methods presented can achieve more reliable probabilistic predictions than existing techniques.

%that the proposed methodology can achieve more reliable probabilistic predictions than existing techniques. In addition, the numerical results show that the proposed confidence intervals can offer informative upper and lower bounds for the actual probabilities that also contain the probabilities predicted by the presented methods.
%than existing techniques.

% %% \vspace{-00.3cm}
\section{Conclusion}
% %% \vspace{-00.2cm}

Reliable probabilistic predictions provided by programmatic weak supervision can significantly improve the cost-efficiency of labeling processes. The new methodology presented in the paper can assess the reliability of probabilistic predictions in terms of confidence intervals with simultaneous coverage probability. In addition, the methods presented result in more reliable probabilistic predictions that minimize the worst-case expected loss. Using multiple benchmark datasets, the numerical results show the performance improvement with respect to the state-of-the-art, and the practicality of the presented confidence intervals for label probabilities. The novel methodology presented can lead to more reliable labeling methods that leverage general types of programmatic weak supervision and provide additional levels of uncertainty awareness.

% \textbf{Limitations :} The confidence intervals presented require to solve two optimization problems per group of instances and label examined. However, this limitation is alleviated by the fact that such problems can be solved very efficiently. To the best of our knowledge, the techniques presented are the first to obtain confidence intervals for probabilistic predictions in programmatic weak supervision that do not require assumptions for the \acp{LF}' behavior. The price to pay for such generality is that the confidence intervals obtained are not tight for small groups of instances or in scenarios with few \acp{LF}.

% \section*{Limitations and Future work}

\textbf{Limitations and future work: }The confidence intervals presented require to solve two optimization problems per group of instances and label examined. However, this limitation is alleviated by the fact that such problems can be solved very efficiently. Future work could explore stochastic gradient descent-based techniques to more efficiently solve the optimization problems for a sizable number of groups.  To the best of our knowledge, the techniques presented are the first to obtain confidence intervals for probabilistic predictions in programmatic weak supervision that do not require assumptions for the LFs' behavior. The price to pay for such generality is that the confidence intervals obtained are not tight for small groups of instances or in scenarios with few LFs. Future work could explore adaptive grouping strategies that can accommodate low-data regimes and identify the need for additional LFs in certain instances. % In addition, in the current version of the paper we propose to obtain groups of instances based on heuristics, as the reviewer points out in the ``Meta'' comment. Future work could explore more sophisticated grouping strategies that better identify specific issues, such as challenging instances for labeling and the need for specific LFs for certain instances.

\begin{appendices}

\section{Strong duality lemma}\label{ap-lemma}

Some of the proofs of the results in the paper make use of Lagrange duality for linear optimization problems over probability distributions. The next lemma provides the duality result needed for such proofs.

\begin{lemma}\label{lem}
Let $\set{U}$ be a non-empty uncertainty set given by \eqref{eq:uncertainty set}. For any function \mbox{$w:\{1,2,\ldots,n\}\times\set{Y}\to\mathbb{R}$}, we have that
\begin{align}\label{duality}
\max_{(\up{p}_1,\up{p}_2,\ldots,\up{p}_n)\in\set{U}} & \frac{1}{n}\sum_{i=1}^n\mathbb{E}_{\up{p}_i}w(i,y)=\min_{\B{\mu}} -\widehat{\B{\tau}}^\top\B{\mu}+\B{\lambda}^\top|\B{\mu}|\\
& +\frac{1}{n}\sum_{i=1}^n\underset{y\in\set{Y}}{\max}\  \big\{\Phi(\Lambda_i,y)^\top\B{\mu}+w(i,y)\big\}.\nonumber
\end{align}
\end{lemma}
\begin{proof}
The optimization problem in the left-hand-side of \eqref{duality} can be written as
\begin{align}\label{opt11}
& \underset{(\up{p}_1,\up{p}_2,\ldots,\up{p}_n)\in\set{U}}{\max} \frac{1}{n}\sum_{i=1}^n\mathbb{E}_{\up{p}_i}w(i,y)\\
& \begin{array}{ccl}=&\underset{\V{p}_1,\V{p}_2,\ldots,\V{p}_n}{\max}&\frac{1}{n}\sum_{i=1}^n\V{p}_i^\top\V{w}_i-\sum_{i=1}^n I_+(\V{p_i})\vspace{0.2cm}\\
&\mbox{s.t.}&\widehat{\B{\tau}}-\B{\lambda}\preceq\frac{1}{n}\sum_{i=1}^n\V{p}_i^\top\Phi(\Lambda_i,\cdot)\preceq\widehat{\B{\tau}}+\B{\lambda}\end{array}\nonumber\end{align}
where $\V{p}_i\in\mathbb{R}^{|\set{Y}|}$ and $\V{w}_i\in\mathbb{R}^{|\set{Y}|}$ for $i=1,2,\ldots,n$ denote the vectors given by the probability distributions $\{\up{p}_i\}$ and functions $\{w(i,\cdot)\}$, and function $I_+$ is given by
$$I_+(\V{p})=\left\{\begin{array}{cc}0&\mbox{if }\V{p}\succeq\V{0}\mbox{ and }\V{p}^\top\V{1}=1 \\\infty&\mbox{otherwise.}\end{array}\right.$$

Optimization problem \eqref{opt11} has Lagrange dual
$$\begin{array}{ccl}&\underset{\B{\mu}_1,\B{\mu}_2}{\min} &-\big(\widehat{\B{\tau}}-\B{\lambda}\big)^\top\B{\mu}_1
+\big(\widehat{\B{\tau}}+\B{\lambda}\big)^\top\B{\mu}_2\\
& &+f^*(\B{\Phi}_1\frac{(\B{\mu}_1-\B{\mu}_2)}{n},\B{\Phi}_2\frac{(\B{\mu}_1-\B{\mu}_2)}{n},\ldots,\B{\Phi}_n\frac{(\B{\mu}_1-\B{\mu}_2)}{n})\\&
\mbox{s.t.}&\B{\mu}_1,\B{\mu}_2\succeq\V{0}\end{array}$$
where $\B{\Phi}_i$ for $i=1,2,\ldots,n$ denotes the matrix with rows $\Phi(\Lambda_i,y)^\top$ for $y\in\set{Y}$, and $f^*$ is the conjugate function of $$f(\V{p}_1,\V{p}_2,\ldots,\V{p}_n)=-\frac{1}{n}\sum_{i=1}^n\V{p}_i^\top\V{w}_i+\sum_{i=1}^nI_+(\V{p}_i)$$
for $\V{w}_i=[w(i,1),w(i,1),\ldots,w(i,T)]^{\top}$. Hence, we have that
\begin{align*}
    f^*(\V{u}_1,\V{u}_2,\ldots,\V{u}_n)=&\sup_{(\up{p}_1,\up{p}_2,\ldots,\up{p}_n)\in\Delta(\set{Y})}\sum_{i=1}^n\V{p}_i^\top\V{u}_i\\
    & +\frac{1}{n}\sum_{i=1}^n\V{p}_i^\top\V{w}_i=\sum_{i=1}^n\big\|\V{u}_i+\frac{1}{n}\V{w}_i\big\|_\infty.
\end{align*}

Therefore, the Lagrange dual above becomes
$$\begin{array}{ccl}&\underset{\B{\mu}_1,\B{\mu}_2}{\min} &-\big(\widehat{\B{\tau}}-\B{\lambda}\big)^\top\B{\mu}_1+\big(\widehat{\B{\tau}}+\B{\lambda}\big)^\top\B{\mu}_2\\
&&+\frac{1}{n}\sum_{i=1}^n\underset{y\in\set{Y}}{\max}\{\Phi(\Lambda_i,y)^\top(\B{\mu}_1-\B{\mu}_2)+w(i,y)\}\\&
\mbox{s.t.}&\B{\mu}_1,\B{\mu}_2\succeq\V{0}.\end{array}$$
It is easy to see that the solution of such optimization problem $\bar{\B{\mu}}_1,\bar{\B{\mu}}_2$ satisfies that $\mbox{$\bar{\mu}_{1}^{(i)}\bar{\mu}_{2}^{(i)}=0$}$ for any $i$ such that $\lambda^{(i)}>0$. Then $\B{\lambda}^\top(\bar{\B{\mu}}_1+\bar{\B{\mu}}_2)=\B{\lambda}^\top|\bar{\B{\mu}}_1-\bar{\B{\mu}}_2|$ and taking $\B{\mu}=\B{\mu}_1-\B{\mu}_2$ the Lagrange dual above is equivalent to the right-hand-side of \eqref{duality}.

Then, the result is obtained because strong duality is satisfied since the constraints in \eqref{opt11} are affine and $\set{U}$ is non-empty. 

\end{proof}
\section{Proof of Theorem~\ref{th:confidence_interval}}\label{app:th_confidence_interval}
\begin{proof}
Let $\set{U}^*$ be the uncertainty set given by \eqref{eq:smallest_uncertainty_set}. % for $\widehat{\B{\tau}}=\B{\tau}=\frac{1}{n}\sum_{i = 1}^n \mathbb{E}_{\up{p}_i^*} \{\Phi(\Lambda_i, y)\}$ and $\B{\lambda}=\V{0}$. 
Such uncertainty is non-empy since it includes the distributions given by actual labels $\mathbb{I}\{y=y_1\},\mathbb{I}\{y=y_2\},\ldots,\mathbb{I}\{y=y_n\}\in\Delta(\set{Y})$.

%Such uncertainty is non-empty and includes the underlying distributions $(\up{p}_1^*,\up{p}_2^*,\ldots,\up{p}_n^*)$. 

For each $\set{I}\subseteq\{1,2,\ldots,n\}$ and $y\in\set{Y}$, using Lemma~\ref{lem} with uncertainty set $\set{U}^*$ and function $w:\{1,2,\ldots,n\}\times\set{Y}\to\mathbb{R}$ given by 
$$w(i,\tilde{y})=\frac{n}{|\set{I}|}\mathbb{I}\{i\in\set{I},\tilde{y}=y\}$$ we have that
\begin{align*}\up{p}_\set{I}^*(y)& =\frac{1}{n}\sum_{i=1}^n w(i,y_i)\\
&\leq\underset{(\up{p}_1,\up{p}_2,\ldots,\up{p}_n)\in\set{U}^*}{\max}\frac{1}{n}\sum_{i=1}^n\sum_{\tilde{y}\in\set{Y}}\up{p}_i(\tilde{y})w(i,\tilde{y})\\
&=\min_{\B{\mu}}  -\B{\tau}^\top\B{\mu}+\frac{1}{n}\sum_{i=1}^n\underset{\tilde{y}\in\set{Y}}{\max}\  \big\{\Phi(\Lambda_i,\tilde{y})^\top\B{\mu}+w(i,\tilde{y})\big\}\\
&\leq -\B{\tau}^\top\overline{\B{\mu}}+\frac{1}{n}\sum_{i=1}^n\underset{\tilde{y}\in\set{Y}}{\max}\  \big\{\Phi(\Lambda_i,\tilde{y})^\top\overline{\B{\mu}}+w(i,\tilde{y})\big\}\\
&=\overline{\up{p}}_\set{I}(y)+(\widehat{\B{\tau}}-\B{\tau})^\top\overline{\B{\mu}}-\B{\lambda}^\top|\overline{\B{\mu}}|
\end{align*}
that directly gives the upper bound of the confidence interval in~\eqref{eq:p1}.

For the lower bound of the confidence interval in \eqref{eq:p1} we use again Lemma~\ref{lem} with uncertainty set $\set{U}^*$ and function
$w:\{1,2,\ldots,n\}\times\set{Y}\to\mathbb{R}$ given by $$w(i,\tilde{y})=-\frac{n}{|\set{I}|}\mathbb{I}\{i\in\set{I},\tilde{y}=y\}.$$ 
Then, we have that 
\begin{align*}\up{p}_\set{I}^*(y) =&-\frac{1}{n}\sum_{i=1}^n w(i,y_i)\\
\geq&\underset{(\up{p}_1,\up{p}_2,\ldots,\up{p}_n)\in\set{U}^*}{\min}-\frac{1}{n}\sum_{i=1}^n\sum_{\tilde{y}\in\set{Y}}\up{p}_i(\tilde{y})w(i,\tilde{y})\\
=&-\underset{(\up{p}_1,\up{p}_2,\ldots,\up{p}_n)\in\set{U}^*}{\max}\frac{1}{n}\sum_{i=1}^n\sum_{\tilde{y}\in\set{Y}}\up{p}_i(\tilde{y})w(i,\tilde{y})\\
=&-\min_{\B{\mu}}  -\B{\tau}^\top\B{\mu}\\
&+\frac{1}{n}\sum_{i=1}^n\underset{\tilde{y}\in\set{Y}}{\max}\  \big\{\Phi(\Lambda_i,\tilde{y})^\top\B{\mu}+w(i,\tilde{y})\big\}\\
=&\max_{\B{\mu}}  \B{\tau}^\top\B{\mu}-\frac{1}{n}\sum_{i=1}^n\underset{\tilde{y}\in\set{Y}}{\max}\  \big\{\Phi(\Lambda_i,\tilde{y})^\top\B{\mu}+w(i,\tilde{y})\big\}\\
=&\max_{\B{\mu}}  -\B{\tau}^\top\B{\mu}+\frac{1}{n}\sum_{i=1}^n\underset{\tilde{y}\in\set{Y}}{\min}\  \big\{\Phi(\Lambda_i,\tilde{y})^\top\B{\mu}-w(i,\tilde{y})\big\}\\
\geq& -\B{\tau}^\top\underline{\B{\mu}}\\
&+\frac{1}{n}\sum_{i=1}^n\underset{\tilde{y}\in\set{Y}}{\min}\  \big\{\Phi(\Lambda_i,\tilde{y})^\top\underline{\B{\mu}}+\frac{n}{|\set{I}|}\mathbb{I}\{i\in\set{I},\tilde{y}=y\}\big\}\\
=&\underline{\up{p}}_\set{I}(y)+(\widehat{\B{\tau}}-\B{\tau})^\top\underline{\B{\mu}}+\B{\lambda}^\top|\overline{\B{\mu}}|
\end{align*}
that directly gives the lower bound of the confidence interval in~\eqref{eq:p1}.

Then, the result in \eqref{eq:actual_prob_interval} is obtained as a direct consequence of \eqref{eq:p1}, and the fact that \mbox{$\mathbb{P}\{|\B{\tau} - \widehat{\B{\tau}}| \preceq \B{\lambda}\} \geq 1 - \delta$.}
\end{proof}

\section{Additional numerical results}\label{app:num}

\begin{figure}
\centering
\psfrag{Probability}[][][0.8]{Probability}
\psfrag{x}[r][r][0.7]{r}
\psfrag{data40}[l][l][0.8]{MMP $y = 0$}% Pr. prediction $y = 0$}
\psfrag{data15}[l][l][0.8]{MMP $y = 1$}% Pr. prediction $y = 1$}
\psfrag{data13}[l][l][0.8]{MMP $y = 0$}% Pr. prediction $y = 0$}
\psfrag{data33}[l][l][0.8]{MMP $y = 1$}% Pr. prediction $y = 1$}
\psfrag{data12}[l][l][0.8]{Actual probability}
\psfrag{data10}[l][l][0.8]{MMP $y = 1$}% Pr. prediction $y = 1$}
\psfrag{data20}[l][l][0.8]{MMP $y = 0$}% Pr. prediction $y = 0$}
\psfrag{data39}[l][l][0.8]{MMP $y = 1$}% Pr. prediction $y = 1$}
\psfrag{data1}[l][l][0.8]{MMP $y = 0$}% Pr. prediction $y = 0$}
\psfrag{data43}[l][l][0.8]{Actual probability}
\psfrag{data19}[l][l][0.8]{Actual probability}
\psfrag{Confidence intervalabc}[l][l][0.8]{Confidence interval}
\psfrag{data2}[l][l][0.8]{MMP $y = 1$}% Confidence interval}
\psfrag{Confidence intervalabcdefghijklmnopq}[l][l][0.8]{Confidence interval $\delta = 0.05$}
\psfrag{data17}[l][l][0.8]{Actual probability}
\psfrag{data3}[l][l][0.8]{Confidence interval $\delta = 0.05$}% Actual probability}
\psfrag{data4}[l][l][0.8]{Confidence interval}
\psfrag{data5}[l][l][0.8]{Actual probability}
\psfrag{r}[l][l][0.8]{$p$}
\psfrag{p}[l][l][0.8]{$p$}
\psfrag{3}[r][r][0.7]{28/36}
\psfrag{7}[r][r][0.7]{21/36}
\psfrag{11}[r][r][0.7]{13/36}
\psfrag{19}[r][r][0.7]{12/36}
\psfrag{23}[r][r][0.7]{9/38}
\psfrag{26}[r][r][0.7]{28/36}
\psfrag{30}[r][r][0.7]{24/36}
\psfrag{34}[r][r][0.7]{20/36}
\psfrag{42}[r][r][0.7]{17/36}
\psfrag{46}[r][r][0.7]{9/36}
\psfrag{1}[r][r][0.7]{.95}
\psfrag{5}[r][r][0.7]{.75}
\psfrag{10}[r][r][0.7]{.90}
\psfrag{14}[r][r][0.7]{.70}
\psfrag{18}[r][r][0.7]{.50}
\psfrag{15}[r][r][0.7]{.75}
\psfrag{20}[r][r][0.7]{.90}
\psfrag{0.4}[r][r][0.7]{.4}
\psfrag{0.6}[r][r][0.7]{.6}
\psfrag{0.8}[r][r][0.7]{.8}
\psfrag{0.7}[r][r][0.7]{.7}
\psfrag{0.85}[r][r][0.7]{.85}
\psfrag{0.9}[r][r][0.7]{.9}
\psfrag{0.95}[r][r][0.7]{.95}
\psfrag{2}[r][r][0.7]{.85}
\psfrag{6}[r][r][0.7]{.65}
\psfrag{1}[r][r][0.7]{1}
% \vspace{-0.3cm}
\includegraphics[width=0.48\textwidth]{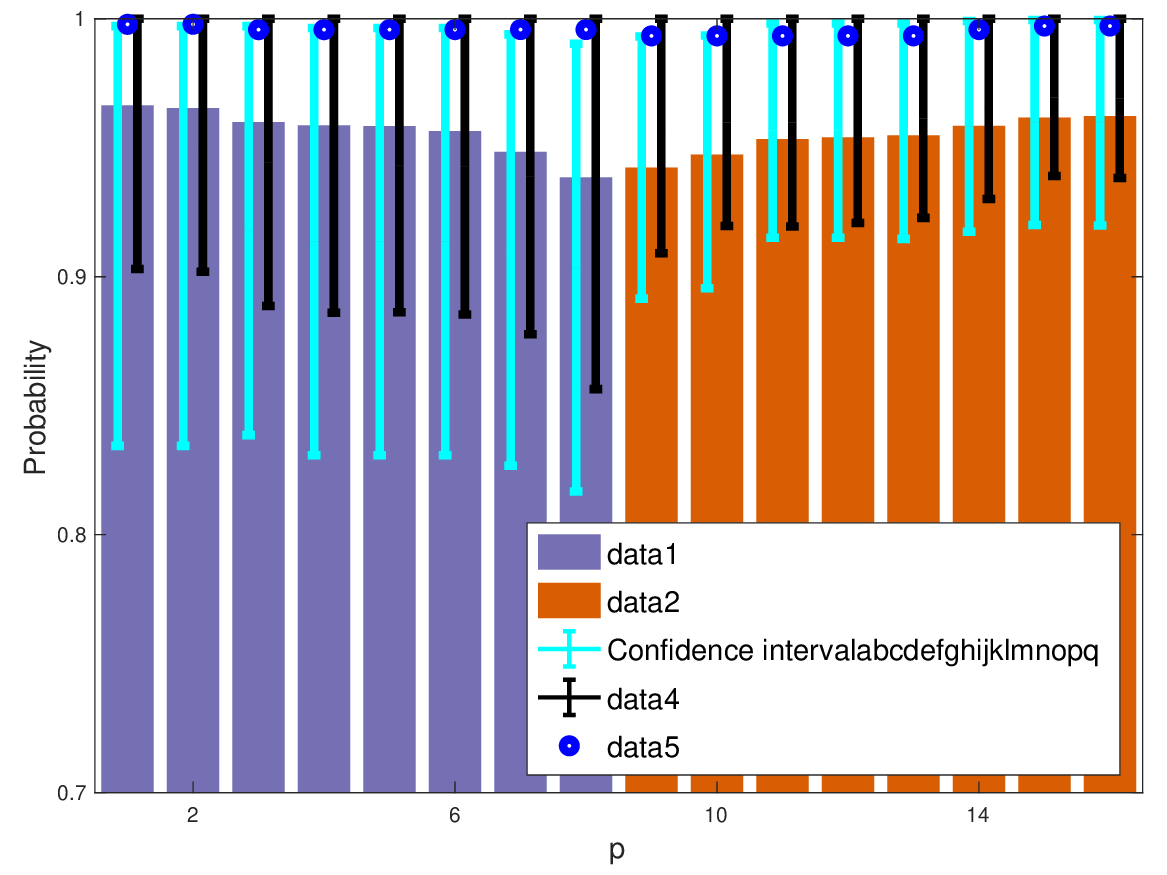}
\caption{% Confidence intervals and MMP probabilities grouping 
Results on IMDB dataset show the reliability of the presented confidence intervals and probabilistic predictions grouping instances as in~\eqref{eq:groups}.}
\label{fig:groups_imdb}
\end{figure}

\begin{figure*}
\centering
\psfrag{class0prob07abcd}[l][l][0.7]{$y = 0$, $p = 0.7$}
\psfrag{data2}[l][l][0.7]{$y = 1$, $p = 0.7$}
\psfrag{data3}[l][l][0.7]{$y = 0$, $p = 0.9$}
\psfrag{data4}[l][l][0.7]{$y = 1$, $p = 0.9$}
\psfrag{Size confidence}[b][][0.8]{Length of confidence interval}
\psfrag{Hyperparam}[t][][0.8]{$1-\B{\delta}$}
\psfrag{0.1}[][][0.7]{.98}
\psfrag{0.10}[][][0.7]{.1}
\psfrag{0.14}[][][0.6]{.14}
\psfrag{0.18}[][][0.6]{.18}
\psfrag{0.22}[][][0.6]{.22}
\psfrag{0.02}[][][0.7]{.9}
\psfrag{0.04}[][][0.7]{.92}
\psfrag{0.06}[][][0.7]{.94}
\psfrag{0.08}[][][0.7]{.96}
\subfloat[Length of confidence intervals using $\B{\lambda}$ given by the Wilson’s interval for a binomial proportion with coverage probability $1 - \delta$ \label{fig:size_conf_lambda_binomial}]{\includegraphics[width=0.48\textwidth]{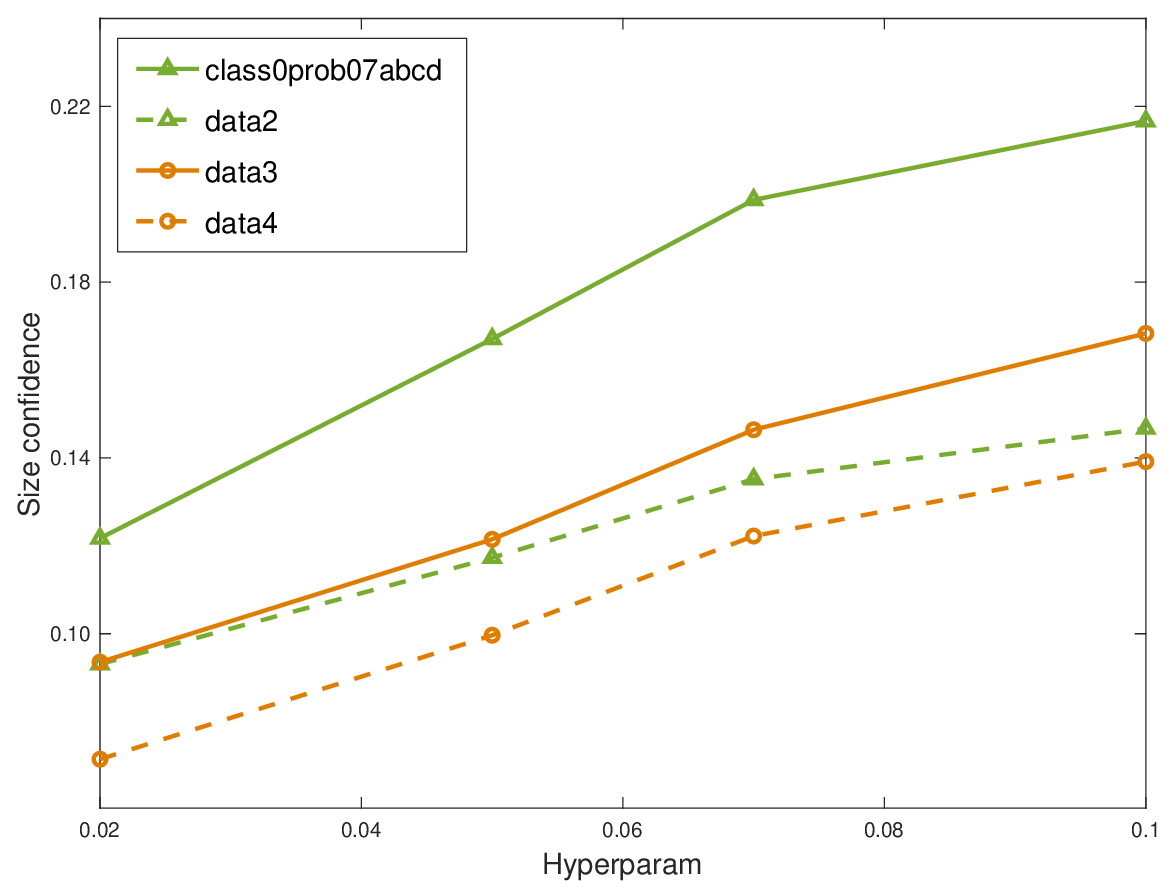}}
\iffalse
             \begin{subfigure}[t]{0.48\textwidth}
         % \subfigure[Class. Error vs \# LFs \label{fig:n_weaks}]{\includegraphics[width=0.48\textwidth]{aa2_nweaks.eps}}\hfill
         {\includegraphics[width=\textwidth]{size_conf_lambda_binomial.eps}}
                  \caption{Length of confidence intervals using $\B{\lambda}$ given by the Wilson’s interval for a binomial proportion with coverage probability $1 - \delta$}
                  \label{fig:size_conf_lambda_binomial}
         \end{subfigure}
         \fi
         \hfill
%         \begin{subfigure}[t]{0.48\textwidth}      
         \psfrag{0.6}[][][0.7]{.6}
         \psfrag{0.7}[][][0.7]{.7}
         \psfrag{0.8}[][][0.7]{.8}
         \psfrag{0.9}[][][0.7]{.9}
         \psfrag{0.1}[][][0.7]{.1}
         \psfrag{1}[][][0.7]{1}
         \psfrag{Hyperparam}[t][][0.8]{$s$}
         \psfrag{class1prob07}[l][l][0.7]{$y = 1$, $p = 0.7$}
\psfrag{class0prob09}[l][l][0.7]{$y = 0$, $p = 0.9$}
\psfrag{c}[l][l][0.7]{$y = 1$, $p = 0.9$}
        % \subfigure[Noisy LFs vs Class. Error \label{fig:proportion_weaks}]
        \subfloat[Length of confidence intervals using $\B{\lambda}$ given by standard sample errors $s/\sqrt{n}$ where $n$ is the number of labeled instances \label{fig:size_conf_lambda}]{\includegraphics[width=0.48\textwidth]{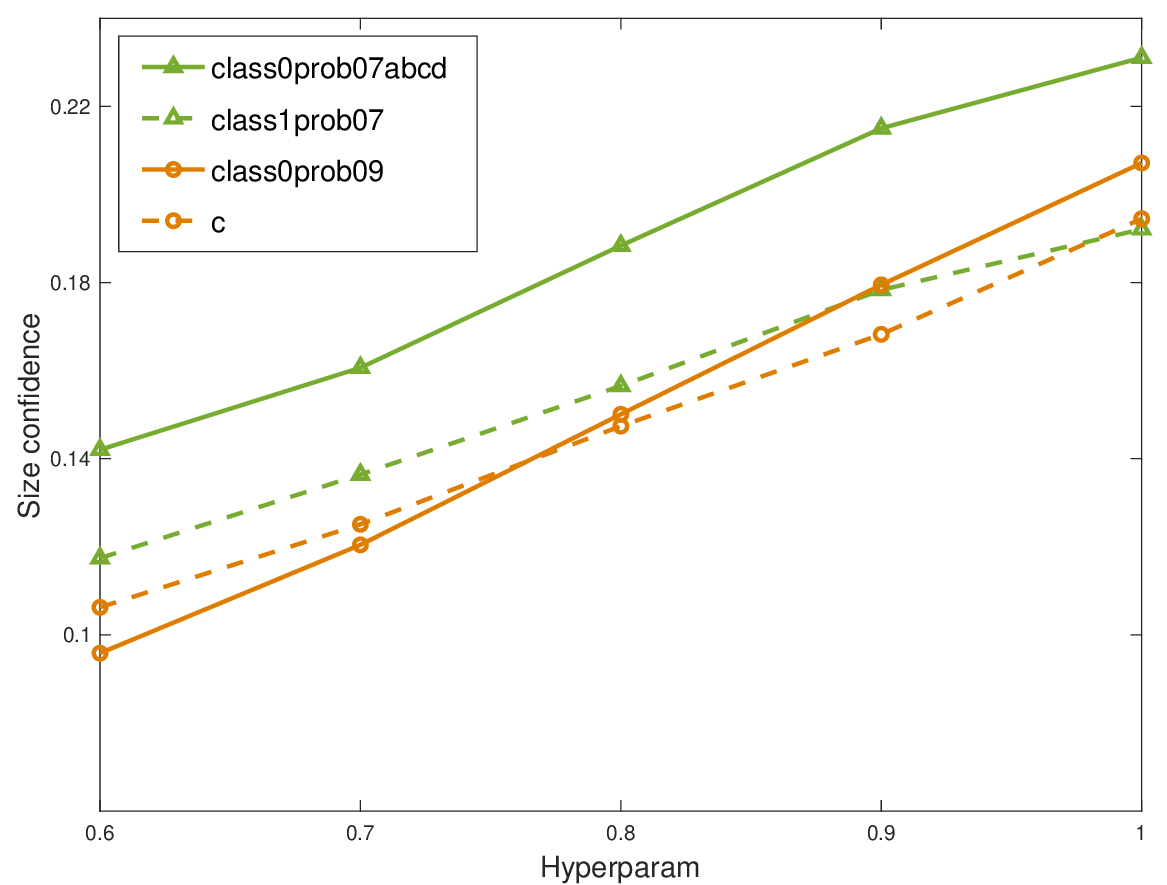}}
        \iffalse
        \includegraphics[width=\textwidth]{size_conf_lambda_std.eps}
                \caption{Length of confidence intervals using $\B{\lambda}$ given by standard sample errors $s/\sqrt{n}$ where $n$ is the number of labeled instances}
                \label{fig:size_conf_lambda}

        \end{subfigure}
                        \fi
        \caption{Results on 'AWA' dataset show the width of the presented confidence intervals.}
 \label{fig:inter_lambda}
\end{figure*}

\begin{figure*}
\centering
               \psfrag{0}[r][r][0.7]{0}
           \psfrag{0.2}[r][r][0.7]{0.2}
          \psfrag{0.6}[r][r][0.7]{0.6}
          \psfrag{0.4}[r][r][0.7]{0.4}
          \psfrag{0.8}[r][r][0.7]{0.8}
          \psfrag{1}[r][r][0.7]{1}
          \psfrag{1.2}[r][r][0.7]{1.2}
          \psfrag{Fraction of positives (positive class 1)}[][t][0.6]{Quantile $q$}
          \psfrag{Mean predicted probability (positive class 1)}[][t][0.6]{Mean predicted probability $C(q)$}
           \psfrag{Snorkel}[l][l][0.8]{Snorkel}
          \psfrag{HyperLM}[l][l][0.8]{Hyper label model}
          \psfrag{EBCC}[l][l][0.8]{EBCC}
          \psfrag{WMRC}[l][l][0.8]{MMP}
          \psfrag{WMRC MV}[l][l][0.8]{MMP-MV}
       \psfrag{Perfectly calibrated abcdefg}[l][l][0.75]{Perfectly calibrated}
         \psfrag{Lower bound, mean, upper bound}[r][r][0.7]{Lower bound, mean, upper bound}
         \psfrag{WMRC unsupervised}[l][l][0.8]{MMP}
         \psfrag{ACML}[l][l][0.8]{AMCL}
                  \psfrag{Snorkel}[l][l][0.8]{Snorkel}
        \psfrag{HyperLMabcdefghijklmn}[l][l][0.8]{Hyper LM}
 \psfrag{EBCC}[l][l][0.8]{EBCC}
  \psfrag{WMRC}[l][l][0.8]{MMP}
  \psfrag{MV}[l][l][0.8]{MV}
        \psfrag{Number of weak classifiers}[][b][0.7]{Number of LFs}
      \psfrag{5}[r][r][0.7]{5}
     \psfrag{15}[r][r][0.7]{15}
      \psfrag{25}[r][r][0.7]{25}
       \psfrag{35}[r][r][0.7]{35}
      \psfrag{0}[r][r][0.7]{0}
       \psfrag{0.04}[r][r][0.7]{.04}
        \psfrag{0.08}[r][r][0.7]{.08}
        \psfrag{0.12}[r][r][0.7]{.12}
        \psfrag{0.16}[r][r][0.7]{.16}
      \psfrag{prediction error}[][t][0.7]{prediction error}
      \psfrag{Classification error}[b][t][0.8]{Prediction error}
         \subfloat[Prediction error varying the number of LFs\label{fig:n_weaks}]{\includegraphics[width=0.48\textwidth]{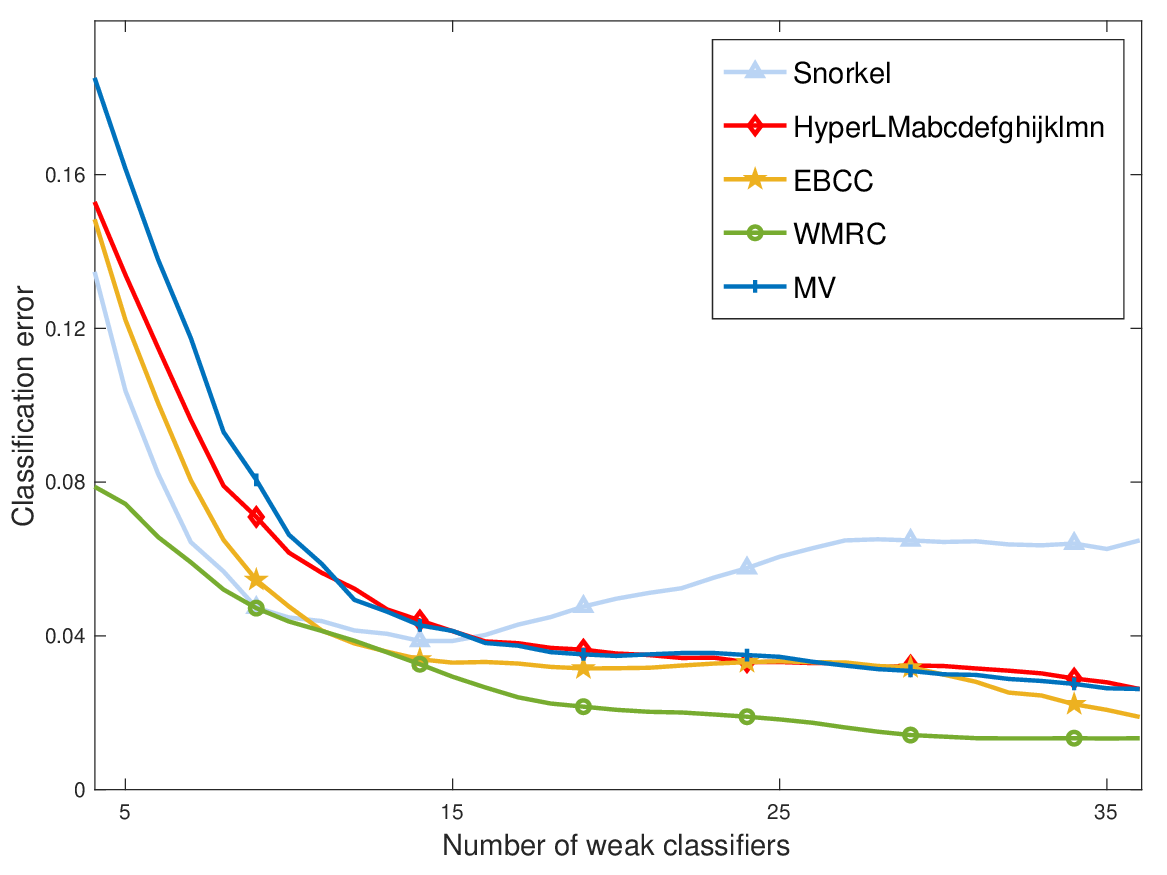}}
         \hfill
         \psfrag{Snorkel}[l][l][0.8]{Snorkel}
         \psfrag{HyperLMabcdefghijklmn}[l][l][0.8]{Hyper LM}
         \psfrag{EBCC}[l][l][0.8]{EBCC}
         \psfrag{WMRC}[l][l][0.8]{MMP}
         \psfrag{MV}[l][l][0.8]{MV}
         \psfrag{Proportion of real weak classifiers}[][b][0.7]{Proportion of LFs from AWA dataset}
         \psfrag{5/36}[r][r][0.7]{5/36}
         \psfrag{15/36}[r][r][0.7]{15/36}
          \psfrag{25/36}[r][r][0.7]{25/36}
         \psfrag{35/36}[r][r][0.7]{35/36}
         \psfrag{0}[r][r][0.7]{0}
          \psfrag{0.2}[r][r][0.7]{.2}
         \psfrag{0.4}[r][r][0.7]{.4}
         \psfrag{0.6}[r][r][0.7]{.6}
         \psfrag{prediction error}[][t][0.6]{prediction error}         
        \subfloat[Robustness to noisy LFs \label{fig:proportion_weaks}]{\includegraphics[width=0.48\textwidth]{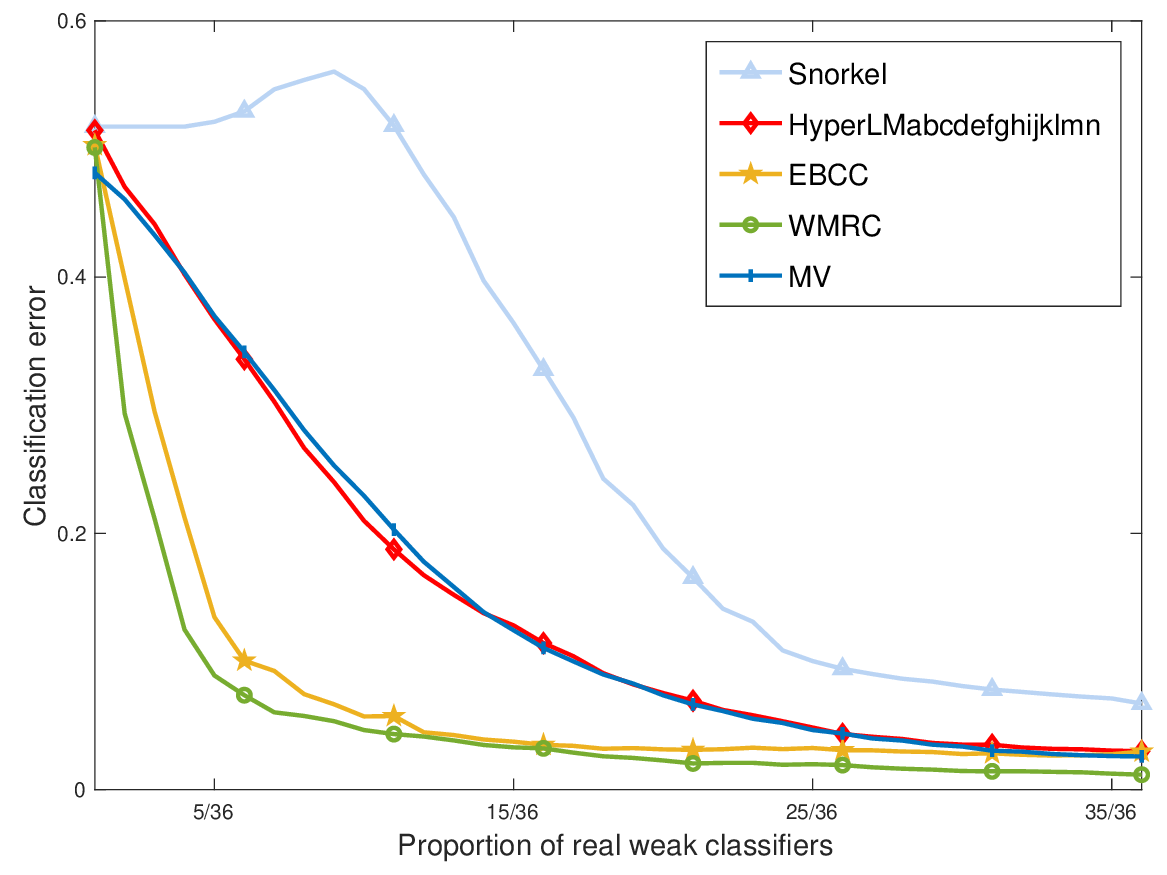}}
        \caption{Results on 'AWA' dataset show the robustness to diverse LFs behavior.}
        \label{fig:LFbehavior}
\end{figure*}

In this section, we include several additional results. In the first set of addition results, we further show the reliability of the presented confidence
intervals and in the second set of additional results, we show the robustness to diverse LFs behaviors. 

In the first set of numerical results, we further illustrate the reliability of the probabilistic predictions and confidence intervals. Specifically, we extend such results using IMDB dataset complementing those in the main paper. Figure~\ref{fig:groups_imdb} shows confidence intervals given by the optimal values of the optimization problems in~\eqref{eq:confidence_interval_upper}-\eqref{eq:confidence_interval_lower} in comparison with the actual probabilities $\up{p}_\set{I}^*(y)$ and probabilistic predictions $\up{h}_\set{I}(y)$ given by MMP methods using IMDB dataset.  This figure shows confidence intervals grouping instances based on the predicted probabilities as in~\eqref{eq:groups} with $p$ given by $\{0.9, 0.85, 0.8, 0.75, 0.7, 0.65, 0.6, 0.55, 0.5\}$. In that figure, we use $\B{\lambda}$ given by the Wilson’s interval for a binomial proportion with coverage probability $1-\delta = 0.95$ (light blue error bar) and $\B{\lambda}$ given by standard sample errors (black error bar). Figure~\ref{fig:groups_imdb} shows that the proposed confidence intervals can provide informative and practically useful bounds for actual label probabilities, even using approximate assessments for estimation errors. In particular, the confidence intervals obtained using standard errors contain the actual and MMP probabilities in all the experimental results carried out.

Figures~\ref{fig:size_conf_lambda_binomial} and \ref{fig:size_conf_lambda} show the length of the confidence intervals for different values of $\B{\lambda}$ in~\eqref{eq:uncertainty set}.  These results are obtained by grouping instances based on the predicted probabilities as in~\eqref{eq:groups} with $p$ given by 0.9 and 0.7. Figure~\ref{fig:size_conf_lambda_binomial} shows confidence intervals using $\B{\lambda}$ given by the Wilson’s interval for a binomial proportion with varying coverage probability $1-\delta$; and Figure~\ref{fig:size_conf_lambda} shows confidence intervals using  $\B{\lambda}$ given by $s/\sqrt{|\set{J}|}$ with $\set{J}$ the set of labeled samples and for different values of~$s$. Such figures show that small values of $\B{\lambda}$ leads to tighter confidence intervals.

  In the second set of numerical results, we show the robustness to diverse LFs behaviors. Figures~\ref{fig:n_weaks} and~\ref{fig:proportion_weaks} show the relationship between the prediction error and the number of LFs and the number of noisy LFs, respectively. These numerical results are obtained averaging, for each number of LFs, the prediction errors achieved with 10 random instantiations in 'AWA' dataset. Figure~\ref{fig:n_weaks} shows the prediction error of the presented method and existing techniques for different number of LFs. Such figure shows the proposed methods significantly improve performance increasing the number of LFs. In addition, Figure~\ref{fig:proportion_weaks} shows the prediction error for different number of LFs from the 'AWA' dataset and generating random LFs up to $36$ LFs. Such random LFs are given by random labels from a Bernoulli distribution with probability $0.5$. Figure~\ref{fig:proportion_weaks} shows that the proposed methods are able to effectively leverage informative LFs and are not affected by the presence of non-informative LFs.

\end{appendices}

\bibliographystyle{IEEEtran}
\bibliography{bibliography} 

\begin{IEEEbiography} 
[{\includegraphics[width=1in, height=1.25in, clip, keepaspectratio]{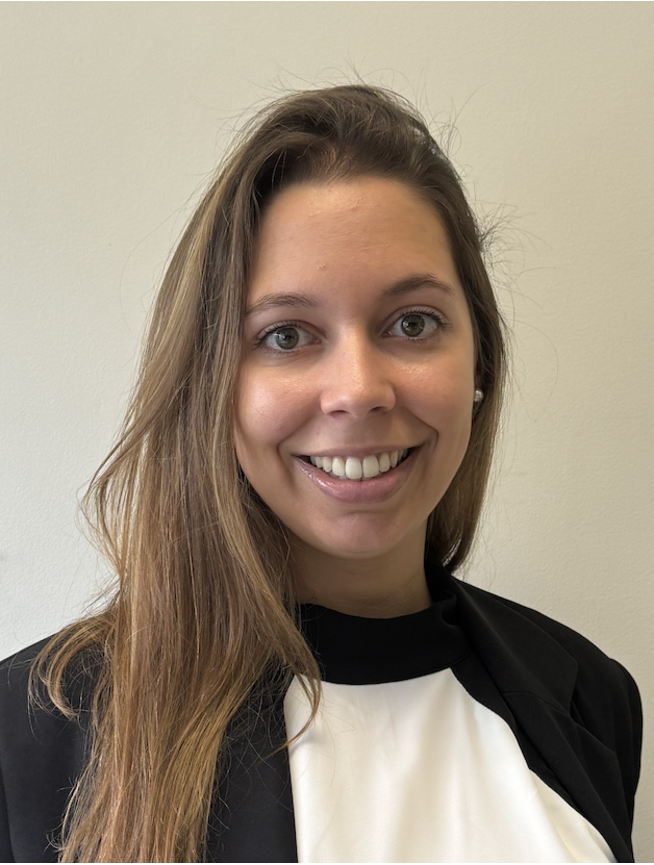}}]{Ver\'onica \'Alvarez} received the Bachelor’s degree in Mathematics from the University of Salamanca, Spain, in 2019 and Ph.D. degree in Computer Engineering from the University of the Basque Country, Spain, in 2023. In 2019, she joined the Machine Learning group at the Basque Center for Applied Mathematics (BCAM) where she worked in the department of Data Science and Machine Learning. From April to July 2023, she was a visiting scholar at the University of California, San Diego (UCSD). She is currently Post-Doctoral Fellow at the Laboratory for Information and Decision Systems (LIDS) in the  Massachusetts Institute of Technology (MIT). 

Verónica has received the Spanish Scientific Society of Computer Science (SCIE)-BBVA Foundation Research Award for Computer Science Young Researchers in 2025.  Her PhD thesis has received the Best Thesis in the Field of Informatics and Artificial Intelligence awarded by the Spanish Society of Artificial Intelligence in 2024, and the Best Doctorate Thesis awarded by the University of the Basque Country in 2024. In addition, she has received the 2022 SEIO-FBBVA Best Applied Contribution in Statistics Field for her work on adaptive probabilistic forecasting, and the Best Paper Award in the 2024 IEEE Sustainable Power and Energy Conference (iSPEC). % She frequently serves as Reviewer for machine learning conferences and journals such as Advances in Neural Information Processing Systems (NeurIPS) and International Conference in Machine Learning (ICML). In addition, her research has been published in prestigious journals and conferences (IEEE Transactions on Power Systems, NeurIPS, ICML, and the Journal for Machine Learning Research) and has garnered significant recognition in the media, including interviews for `El Mundo' and `Antena 3 TV'.
\end{IEEEbiography}

\begin{IEEEbiography} 
[{\vspace{-0.cm}\includegraphics[width=1.05in, height=1.4in, clip, keepaspectratio]{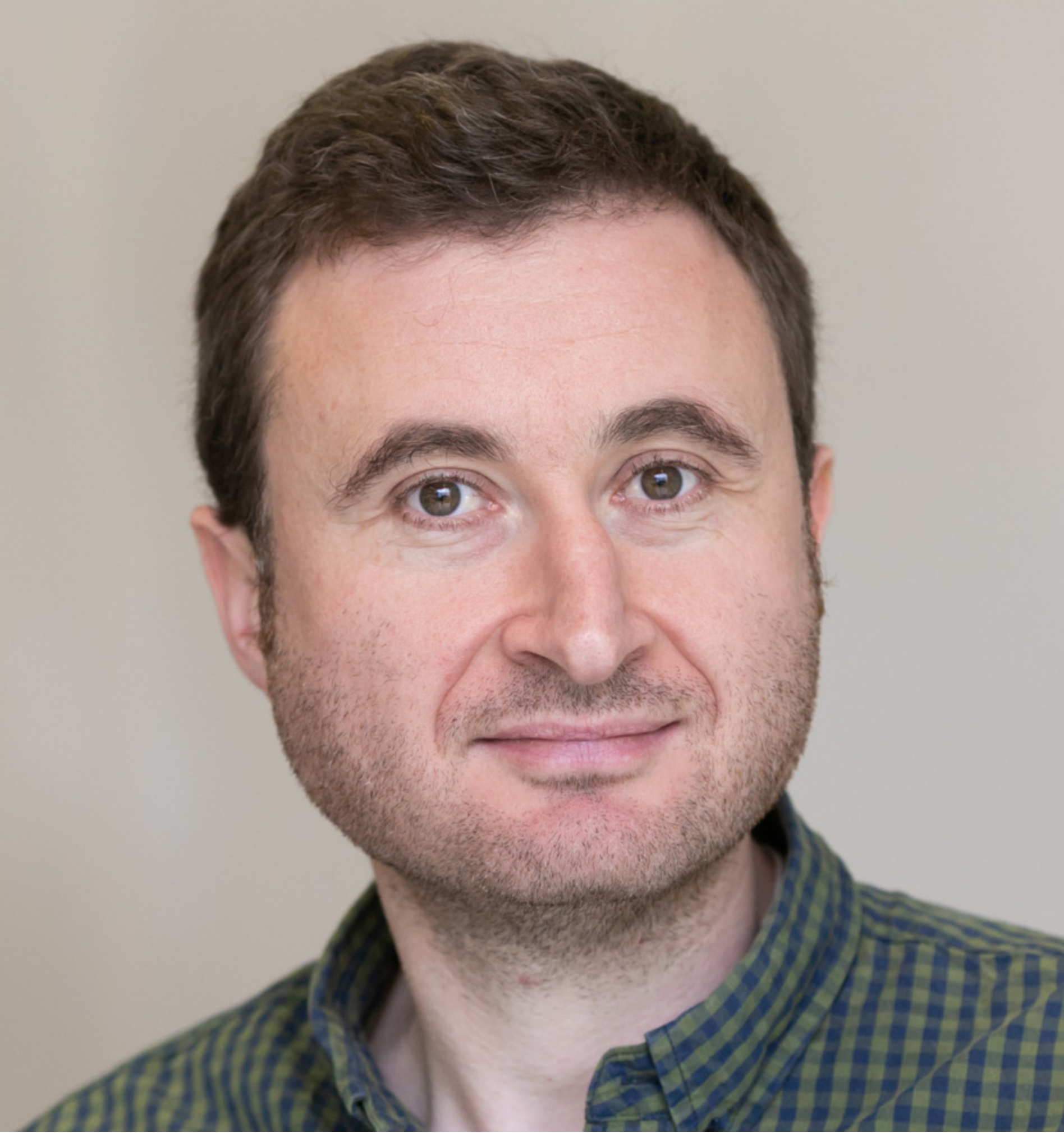}}]
{Santiago Mazuelas}
(Senior Member, IEEE) received the Ph.D. in Mathematics and Ph.D. in Telecommunications Engineering from the University of Valladolid, Spain, in 2009 and 2011, respectively.

Since 2017 he has been with the Basque Center for Applied Mathematics (BCAM) where he is currently Ikerbasque Research Associate. Prior to joining BCAM, he was a Staff Engineer at Qualcomm Corporate Research and Development from 2014 to 2017. He previously worked from 2009 to 2014 as Postdoctoral Fellow and Associate at the Laboratory for Information and Decision Systems (LIDS) at the Massachusetts Institute of Technology (MIT). His current main research interests are on supervised and weakly supervised learning methods for classification problems. 

Dr. Mazuelas is currently Associate Editor-in-Chief for the IEEE TRANSACTIONS ON MOBILE COMPUTING and Associate Editor for the IEEE TRANSACTIONS ON WIRELESS COMMUNICATIONS, and he was Area Editor for the IEEE COMMUNICATIONS LETTERS from 2017 to 2022. He served as Technical Program Vice-chair at the 2021 IEEE Globecom as well as Symposium Co-chair at the 2014 IEEE Globecom, the 2015 IEEE ICC, and the 2020 IEEE ICC. He has received the Young Scientist Prize from the Union Radio-Scientifique Internationale (URSI) Symposium in 2007, and the Early Achievement Award from the IEEE ComSoc in 2018. His papers received the IEEE Communications Society Fred W. Ellersick Prize in 2012, the SEIO-BBVA Foundation Best Contribution in 2022 and 2025,, and Best Paper Awards from the IEEE ICC in 2013, the IEEE ICUWB in 2011, the IEEE Globecom in 2011, and the IEEE iSPEC in 2024.
\end{IEEEbiography} 

\begin{IEEEbiography} 
[{\includegraphics[width=1in, height=1.25in, clip, keepaspectratio]{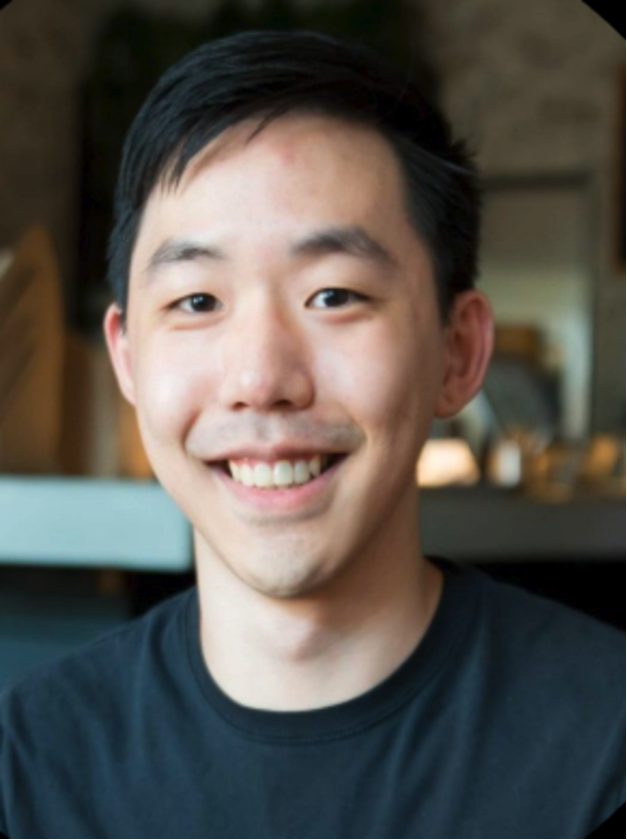}}]{Steven An} received a B.S. degree in Mathematical and Scientific Computation with Honors and an A.B. in Philosophy with Highest Honors from the University of California, Davis in 2019. In 2021, he received an M.S. degree in Computer Science at the University of California, San Diego. He is currently a Ph.D. student at the University of California, San Diego, with a focus on semi/weakly supervised ensemble methods for classification.
\end{IEEEbiography} 

\begin{IEEEbiography} 
[{\includegraphics[width=1in, height=1.25in, clip, keepaspectratio]{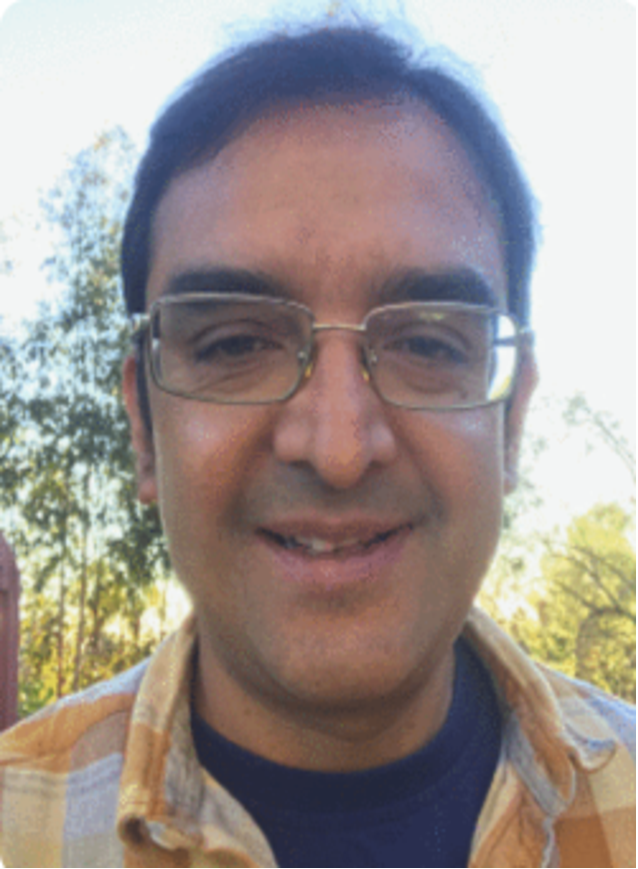}}]{Sanjoy Dasgupta} received the A.B. degree from Harvard University, Cambridge, MA, USA, in 1993 and the Ph.D. degree from the University of California, Berkeley, Berkeley, CA, USA, in 2000, both in computer science.

Prof. Dasgupta spent two years with AT\&T Research Labs, Florham Park, NJ, USA, before joining the University of California, San Diego, La Jolla, CA, USA, where he is currently a Professor of Computer Science and Engineering. His area of research is algorithmic statistics, with a focus on unsupervised and minimally supervised learning. 

Sanjoy is the author of the textbook, Algorithms (with Christos Papadimitriou and Umesh Vazirani), which appeared in 2006. In addition, Prof. Dasgupta was a Program Co-Chair of the Conference on Learning Theory (COLT) in 2009, Program Co-Chair for the International Conference on Machine Learning (ICML) in 2013, and General Chair for the International Conference on Artificial Intelligence and Statistics (AISTATS) in 2024.
\end{IEEEbiography} 

\end{document}